\documentclass[review]{elsarticle}
\usepackage[utf8]{inputenc}
\usepackage{pifont}
\usepackage{tikz}
\usepackage{caption}
\usepackage{subcaption}
\usepackage{url}

\usepackage{amsmath}
\usepackage{amssymb}
\usepackage{amsthm}
\usepackage{booktabs}

\journal{Neural Networks}

\newcommand{\ie}{i.e., }
\newcommand{\eg}{e.g., }
\newcommand{\quotes}[1]{``#1''}

\title{A Gentle Introduction to Deep Learning for Graphs}
\author[unipi]{Davide Bacciu}
\ead{bacciu@di.unipi.it}
\author[unipi]{Federico Errica}
\ead{federico.errica@phd.unipi.it}
\author[unipi]{Alessio Micheli}
\ead{micheli@di.unipi.it}
\author[unipi]{Marco Podda}
\ead{marco.podda@di.unipi.it}
\address[unipi]{Department of Computer Science,
University of Pisa,
Italy.}

\begin{document}

\begin{abstract}
The adaptive processing of graph data is a long-standing research topic that has been lately consolidated as a theme of major interest in the deep learning community. 
The snap increase in the amount and breadth of related research has come at the price of little systematization of knowledge and attention to earlier literature. This work is a tutorial introduction to the field of deep learning for graphs. It favors a consistent and progressive presentation of the main concepts and architectural aspects over an exposition of the most recent literature, for which the reader is referred to available surveys. The paper takes a top-down view of the problem, introducing a generalized formulation of graph representation learning based on a local and iterative approach to structured information processing. Moreover, it introduces the basic building blocks that can be combined to design novel and effective neural models for graphs. We complement the methodological exposition with a discussion of interesting research challenges and applications in the field.
\end{abstract}


\maketitle

\section{Introduction}
Graphs are a powerful tool to represent data that is produced by a variety of artificial and natural processes. A graph has a compositional nature, being a compound of atomic information pieces, and a relational nature, as the links defining its structure denote relationships between the linked entities. Also, graphs allow us to represent a multitude of associations through link orientation and labels, such as discrete relationship types, chemical properties, and strength of the molecular bonds.

But most importantly, graphs are ubiquitous. In chemistry and material sciences, they represent the molecular structure of a compound, protein interaction and drug interaction networks, biological and bio-chemical associations. In social sciences, networks are widely used to represent people's relationships, whereas they model complex buying behaviors in recommender systems. 

The richness of such data, together with the increasing availability of large repositories, has motivated a recent surge in interest in deep learning models that process graphs in an adaptive fashion. The methodological and practical challenges related to such an overarching goal are several. First, learning models for graphs should be able to cater for samples that can vary in size and topology. In addition to that, information about node identity and ordering across multiple samples is rarely available. Also, graphs are discrete objects, which poses restrictions 
to differentiability. Moreover, their combinatorial nature hampers the application of exhaustive search methods. Lastly, the most general classes of graphs allow the presence of loops, which are a source of complexity when it comes to message passing and node visits. In other words, dealing with graph data brings in unparalleled challenges, in terms of expressiveness and computational complexity, when compared to learning with vectorial data. Hence, this is an excellent development and testing field for novel neural network methodologies.

Despite the recent burst of excitement of the deep learning community, the area of neural networks for graphs has a long-standing and consolidated history, rooting in the early nineties with seminal works on Recursive Neural Networks for tree structured data (see \cite{sperduti_supervised_1997, frasconi_general_1998, bianucci_application_2000} and the references therein). Such an approach has later been rediscovered within the context of natural language processing applications \cite{socher_parsing_2011, tai_improved_2015}. Also,
it has been progressively extended to more complex and richer structures, starting from directed acyclic graphs \cite{micheli_contextual_2004}, for which universal approximation guarantees have been given \cite{hammer_universal_2005}. The recursive processing of structures has also been leveraged by probabilistic approaches, first as a purely theoretical model \cite{frasconi_general_1998} and later more practically through efficient approximated distributions \cite{bacciu_compositional_2012}. 

The recursive models share the idea of a (neural) state transition system that traverses the structure to compute its embedding.
The main issue in extending such approaches to general graphs (cyclic/acyclic, directed/undirected) was the processing of cycles. Indeed, the \textit{mutual dependencies} between state variables cannot be easily modeled by the recursive neural units. The earliest models to tackle this problem have been the Graph Neural Network \cite{scarselli_graph_2009} and the Neural Network for Graphs \cite{micheli_neural_2009}. The former is based on a state transition system similar to the recursive neural networks, but it allows cycles in the state computation within a contractive setting of the dynamical system. The Neural Network for Graphs, instead, exploits the idea that mutual dependencies can be managed by leveraging the representations from previous layers in the architecture. This way, the model breaks the recursive dependencies in the graph cycles with a multi-layered architecture. Both models have pioneered the field by laying down the foundations of two of the main approaches for graph processing, namely the \textit{recurrent} \cite{scarselli_graph_2009} and the \textit{feedforward} \cite{micheli_neural_2009} ones. In particular, the latter has now become the predominant approach, under the umbrella term of graph convolutional (neural) networks (named after approaches \cite{kipf_semi-supervised_2017,hamilton_inductive_2017} which reintroduced the above concepts around 2015).

This paper takes pace from this historical perspective to provide a gentle introduction to the field of neural networks for graphs, also referred to as deep learning for graphs in modern terminology. It is intended to be a paper of tutorial nature, favoring a well-founded, consistent, and progressive opening to the main concepts and building blocks to assemble deep architectures for graphs. Therefore, it does not aim at being an exposition of the most recently published works on the topic. The motivations for such a tutorial approach are multifaceted. On the one hand, the surge of recent works on deep learning for graphs has come at the price of a certain forgetfulness, if not lack of appropriate referencing, of pioneering and consolidated works. As a consequence, there is the risk of running through a wave of rediscovery of known results and models. On the other hand, the community is starting to notice troubling trends in the assessment of deep learning models for graphs \cite{shchur_pitfalls_2018,errica_fair_2020}, which calls for a more principled approach to the topic. Lastly, a certain number of survey papers have started to appear in the literature \cite{battaglia_relational_2018, bronstein_geometric_2017,gilmer_neural_2017, hamilton_representation_2017,zhang_deep_2018, wu_comprehensive_2019, zhang_graph_2019}, while a more slowly-paced introduction to the methodology seems lacking. 

This tutorial takes a top-down approach to the field while maintaining a clear historical perspective on the introduction of the main concepts and ideas. To this end, in Section \ref{sec:overview}, we first provide a generalized formulation of the problem of representation learning in graphs, introducing and motivating the architecture roadmap that we will be following throughout the rest of the paper. We will focus, in particular, on methods that deal with local and iterative processing of information as these are more consistent with the operational regime of neural networks. In this respect, we will pay less attention to global approaches (\ie assuming a single fixed adjacency matrix) based on spectral graph theory. We will then proceed, in Section \ref{sec:building-blocks}, to introduce the basic building blocks that can be assembled and combined to create modern deep learning architectures for graphs. In this context, we will introduce the concepts of graph convolutions as local neighborhood aggregation functions, the use of attention, sampling, and pooling operators defined over graphs, and we will conclude with a discussion on aggregation functions that compute whole-structure embeddings. Our characterization of the methodology continues, in Section \ref{sec:learning-criteria}, with a discussion of the main learning tasks undertaken in graph representation learning, together with the associated cost functions and a characterization of the related inductive biases. The final part of the paper surveys other related approaches and tasks (Section \ref{sec:otherworks}), and it discusses interesting research challenges (Section \ref{sec:open}) and applications (Section \ref{sec:applic}). We conclude the paper with some final considerations and hints for future research directions. 
\section{High-level Overview}
\label{sec:overview}
We begin with a high-level overview of deep learning for graphs. To this aim, we first summarize the necessary mathematical notation. Secondly, we present the main ideas the vast majority of works in the literature borrow from.

\subsection{Mathematical Notation}
\begin{figure}[t]
\centering
\begin{subfigure}{.3\textwidth}
  \centering
  \resizebox{0.8\textwidth}{!}{\includegraphics{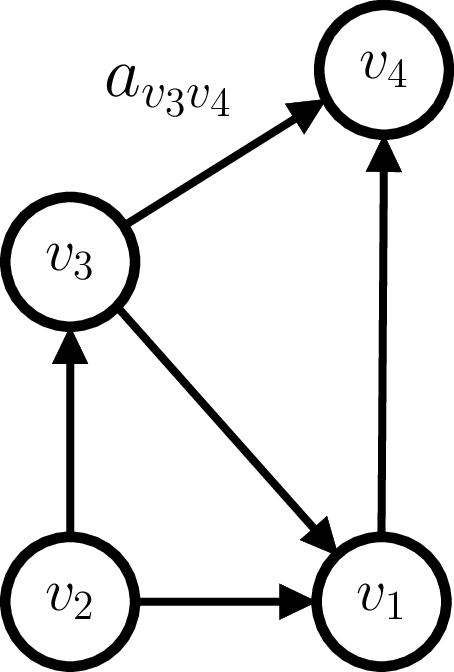}}
  \caption{}
  \label{subfig:graph1}
\end{subfigure}%
\begin{subfigure}{.3\textwidth}
  \centering
   \resizebox{0.85\textwidth}{!}{\includegraphics{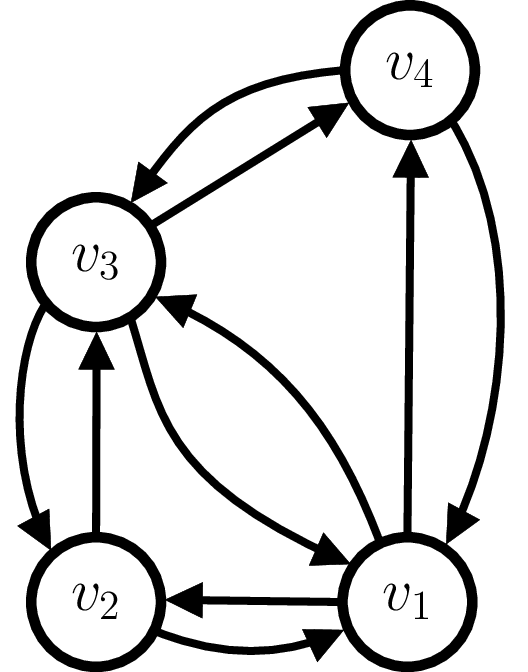}}
  \caption{}
  \label{subfig:graph2}
\end{subfigure}
\begin{subfigure}{.3\textwidth}
  \centering
  \resizebox{0.95\textwidth}{!}{\includegraphics{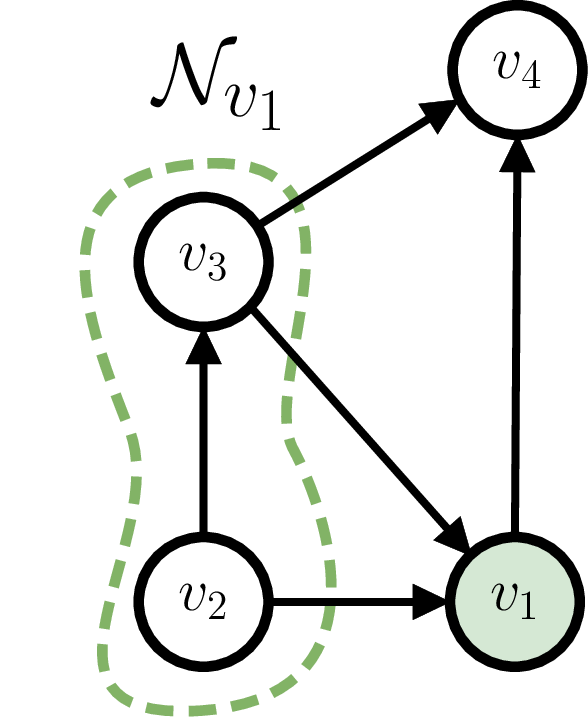}}
  \caption{}
  \label{subfig:graph3}
\end{subfigure}
\caption{(a) A directed graph with oriented arcs is shown. (b) If the graph is undirected, we can transform it into a directed one to obtain a viable input for graph learning methods. In particular, each edge is replaced by two oriented and opposite arcs with identical edge features. (c) We visually represent the (open) neighborhood of node $v_1$.}\label{fig:graph}
\end{figure}

Formally, a graph $g = (\mathcal{V}_g,\mathcal{E}_g,\mathcal{X}_g,\mathcal{A}_g)$ is defined by a set of \emph{vertexes} $\mathcal{V}_g$ (also referred to as \emph{nodes}) and by a set of \emph{ edges} (or \emph{arcs}) $\mathcal{E}_g$ connecting pairs of nodes \citep{bondy_graph_1976}. When the pairs are unordered, \ie $\mathcal{E}_g \subseteq \{ \{u,v\} \mid u, v \in \mathcal{V}_g\}$, we speak of \textit{undirected} graphs and \textit{non-oriented} edges. On the other hand, when pairs of nodes are ordered, \ie $\mathcal{E}_g \subseteq \{ (u,v) \mid u, v \in \mathcal{V}_g\}$, we say a graph is \textit{directed} and its edges are \textit{oriented}. In both cases, the ends of an edge are said to be \textit{incident} with the edge and vice versa. $\mathcal{E}_g$ specifies how nodes are interconnected in the graph, but we can also encode this structural information into an \textit{adjacency matrix}. Specifically, the adjacency matrix of $g$ is the ${|\mathcal{V}_g| \times |\mathcal{V}_g|}$ binary square matrix $\mathbf{A} $ where $\mathbf{A}_{uv} \in \{0,1\}$ is 1 if there is an arc connecting $u$ and $v$, and it is 0 otherwise. It follows that the matrix $\mathbf{A}$ of an undirected graph is symmetric, but the same does not necessarily hold true for a directed graph. Figure \ref{subfig:graph1} depicts a directed graph with oriented arcs.

In many practical applications, it is useful to enrich the graph $g$ with additional node and edge information belonging to the domains $\mathcal{X}_g$ and $\mathcal{A}_g$, respectively. Each node $u \in |\mathcal{V}_g|$ is associated with a particular feature vector $\mathbf{x}_v \in \mathcal{X}_g$, while each edge holds a particular feature vector $\mathbf{a}_{uv} \in \mathcal{A}_g$. In \cite{frasconi_general_1998}, this is referred to as nodes (respectively edges) being \quotes{uniformly labelled}.
In the general case one can consider $\mathcal{X}_g \subseteq \mathbb{R}^d, d \in \mathbb{N}$ and $\mathcal{A}_g \subseteq \mathbb{R}^{d'}, d' \in \mathbb{N}$. Here the terms $d$ and $d’$ denote the number of features associated with each node and edge, respectively. Note that, despite having defined node and edge features on the real set for the sake of generality, in many applications these take discrete values. Moreover, from a practical perspective, we can think of a graph with no node ( respectively edge) features as an equivalent graph in which all node (edge) features are identical.

As far as undirected graphs are concerned, these are straightforwardly transformed to their directed version. In particular, every edge $\{u,v\}$ is replaced by two distinct and oppositely oriented arcs $(u,v)$ and $(v,u)$, with identical edge features as shown in Figure \ref{subfig:graph2} .

A \textit{path} is a sequence of edges that joins a sequence of nodes. Whenever there exists a non-empty path from a node to itself with no other repeated nodes, we say the graph has a \textit{cycle}; when there are no cycles in the graph, the graph is called \textit{acyclic}.

A topological ordering of a directed graph $g$ is a total sorting of its nodes such that for every directed edge $(u,v)$ from node $u$ to node $v$, $u$ comes before $v$ in the ordering. A topological ordering exists if and only if the directed graph has no cycles, \ie if it is a directed acyclic graph (DAG).

A graph is \textit{ordered} if, for each node $v$, a total order on the edges incident on $v$ is defined and \textit{unordered} otherwise. Moreover, a graph is \textit{positional} if, besides being ordered, a distinctive positive integer is associated with each edge incident on a node $v$ (allowing some positions to be absent) and \emph{non-positional} otherwise. To summarize, in the rest of the paper we will assume a general class of directed/undirected, acyclic/cyclic and positional/non-positional graphs.

The neighborhood of a node $v$ is defined as the set of nodes which are connected to $v$ with an oriented arc, \ie $\mathcal{N}_v = \{ u \in \mathcal{V}_g | (u,v) \in \mathcal{E}_g\}$. $\mathcal{N}_v$ is \textit{closed} if it \textit{always} includes $u$ and \textit{open} otherwise. If the domain of arc labels $\mathcal{A}$ is discrete and finite, \ie $\mathcal{A} =  \{c_1, \ldots, c_m\}$, we define the subset of neighbors of $v$ with arc label $c_k$ as $\mathcal{N}_v^{c_k} =\{u \in \mathcal{N}_v \mid \mathbf{a}_{uv} = c_k\}$. Figure \ref{subfig:graph3} provides a graphical depiction of the (open) neighborhood of node $v_1$.

In supervised learning applications, we may want to predict an output for a single node, an edge, or an entire graph, whose ground truth labels are referred to as $y_v$, $y_{uv}$, and $y_g$ respectively. Finally, when clear from the context, we will omit the subscript $g$ to avoid verbose notation.

\subsection{The Bigger Picture}\label{sec:big-picture}

\begin{figure}[t]
\centering
\resizebox{\textwidth}{!}{\includegraphics{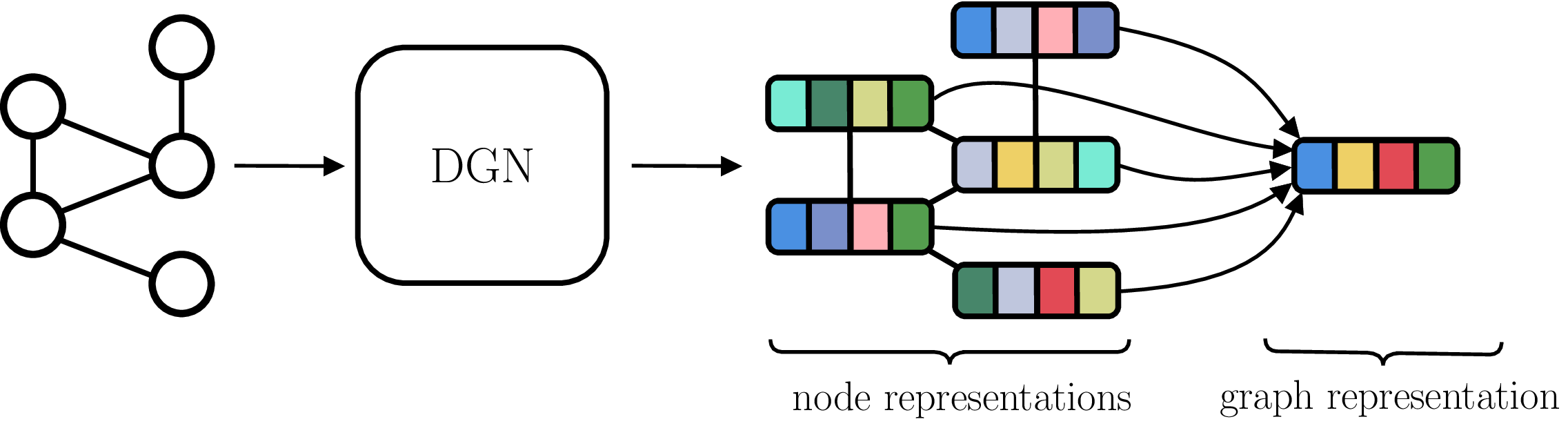}}
\caption{The bigger picture that all graph learning methods share. A \quotes{Deep Graph Network} takes an input graph and produces node representations $\mathbf{h}_v \ \forall v \in \mathcal{V}_g$. Such representations can be aggregated to form a single graph representation $\mathbf{h}_g$.}\label{fig:bigger-picture}
\end{figure}

Regardless of the training objective one cares about, almost all deep learning models working on graphs ultimately produce node representations, also called states. The overall mechanism is sketched in Figure \ref{fig:bigger-picture}, where the input graph on the left is mapped by a model into a graph of node states with the same topology. In \citep{frasconi_general_1998}, this process is referred to as performing an \textit{isomorphic transduction} of the graph. This is extremely useful as it allows tackling nodes, edges, and graph-related tasks. For instance, a graph representation can be easily computed by aggregating together its nodes representations, as shown in the right-hand side of Figure \ref{fig:bigger-picture}.

To be more precise, each node in the graph will be associated with a state vector $\mathbf{h}_v \ \forall v \in \mathcal{V}_g$. The models discussed in this work visit/traverse the input graph to compute node states. Importantly, in our context of general graphs, the result of this traversal does not depend on the visiting order and, in particular, no topological ordering among nodes is assumed. Being independent of a topological ordering has repercussions on how deep learning models for graphs deal with cycles (Section 2.3). Equivalently, we can say that the state vectors can be computed by the model in parallel for each node of the input graph.

\begin{figure}[htb]
\centering
\resizebox{\textwidth}{!}{\includegraphics{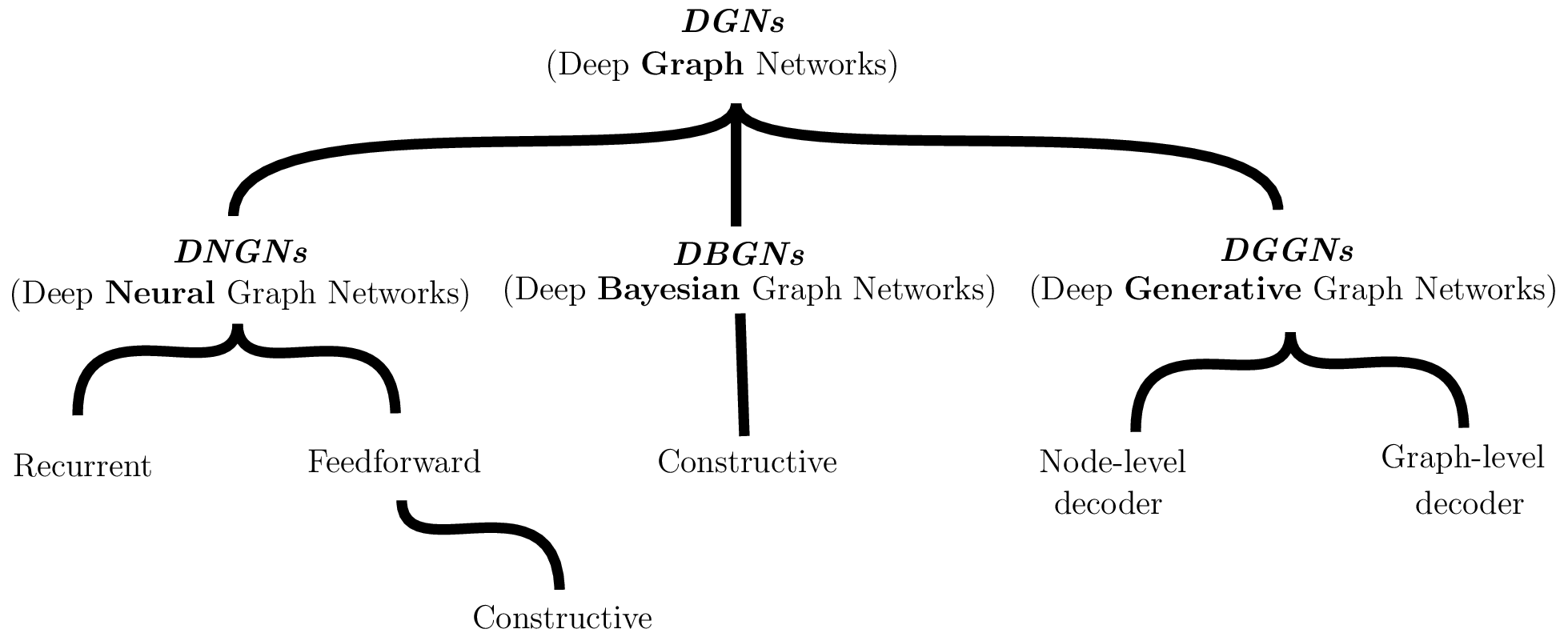}}
\caption{The road-map of the architectures we will discuss in detail.}\label{fig:taxonomy}
\end{figure}

The work of researchers and practitioners therefore revolves around the definition of deep learning models that automatically extract the relevant features from a graph. In this tutorial, we refer to such models with the uniforming name of \quotes{Deep Graph Networks} (DGNs). On the one hand, this general terminology serves the purpose of disambiguating the terms \quotes{Graph Neural Network}, which we use to refer to \cite{scarselli_graph_2009}, and \quotes{Graph Convolutional Network}, which refers  to, \eg \cite{kipf_semi-supervised_2017}. These two terms have been often used across the literature to represent the whole class of neural networks operating on graph data, generating ambiguities and confusion among practitioners. On the other hand, we also use it as the base of an illustrative taxonomy (shown in Figure \ref{fig:taxonomy}), which will serve as a road-map of the discussion in this and the following sections. \\ 
Note that with the term \quotes{DGN} (and its taxonomy) we would like to focus solely on the part of the deep learning model that learns to produce node representations. Therefore, the term does not encompass those parts of the architecture that compute a prediction, \eg the output layer. In doing so, we keep a modular view on the architecture, and we can combine a deep graph network with any predictor that solves a specific task.

We divide deep graph networks into three broad categories. The first is called Deep Neural Graph Networks (DNGNs), which includes models inspired by neural architectures. The second category is that of Deep Bayesian Graph Networks (DBGNs), whose representatives are probabilistic models of graphs. Lastly, the family of Deep Generative Graph Networks (DGGNs) leverages both neural and probabilistic models to generate graphs. This taxonomy is by no means a strict compartmentalization of methodologies; in fact, all the approaches we will focus on in this tutorial are based on local relations and iterative processing to diffuse information across the graph, regardless of their neural or probabilistic nature. 

\subsection{Local relations and iterative processing of information}
Learning from a population of arbitrary graphs raises two fundamental issues: i) no assumptions about the topology of the graph hold in general, and ii) structures may contain cycles. We now discuss both points, highlighting the most common solutions adopted in the literature.

\paragraph{Graphs with variable topology} First of all, we need a way to seamlessly process information of graphs that vary both in size and shape. In the literature, this has been solved by building models that work \textit{locally} at node level rather than at graph level. In other words, the models process each node using information coming from the neighborhood. This recalls the localized processing of images in convolutional models \cite{lecun_convolutional_1995}, where the focus is on a single pixel and its set of finite neighbors (however defined). Such \textit{stationarity} assumption allows reducing significantly the number of parameters needed by the model, as they are re-used across all nodes (similarly to how convolutional filters are shared across pixels). Moreover, it effectively and efficiently combines the \quotes{experience} of all nodes and graphs in the dataset to learn a single function. At the same time, the stationarity assumption calls for the introduction of mechanisms that can learn from the global structure of the graph as well, which we discuss in the following section.

Notwithstanding these advantages, local processing alone does not solve the problem of graphs of variable neighborhood shape. This issue arises in the case of non-positional graphs, where there is no consistent way to order the nodes of a neighborhood. In this case, one common solution is to use permutation invariant functions acting on the neighborhood of each node. A permutation invariant function is a function whose output does not change upon reordering of the input elements. Thanks to this property, these functions are well suited to handle an arbitrary number of input elements, which comes in handy when working on unordered and non-positional graphs of variable topology. Common examples of such functions are the sum, mean, and product of the input elements.
Under some conditions, it is possible to approximate all permutation invariant continuous functions by means of suitable transformations \cite{zaheer_deep_2017, wagstaff_limitations_2019}. More concretely, if the input elements belong to an uncountable space $\mathcal{X}$, \eg $\mathbb{R}^d$, and they are in finite and fixed number $M$, then any permutation invariant continuous function $\Psi: \mathcal{X}^M \rightarrow \mathcal{Y}$ can be expressed as (Theorem 4.1 of \citep{wagstaff_limitations_2019})
\begin{align}
\Psi(Z) = \phi(\sum_{z \in Z} \psi(z)),
\end{align}
where $\phi: M \rightarrow \mathcal{Y}$ and $\psi: \mathcal{X} \rightarrow M$ are continuous functions such as neural networks (for the universal approximation theorem \citep{cybenko_approximation_1989}). Throughout the rest of this work, we will use the Greek letter $\Psi$ to denote permutation invariant functions.

\paragraph{Graphs contain cycles} A graph cycle models the presence of mutual dependencies/influences between the nodes. In addition, the local processing of graphs implies that any intermediate node state is a function of the state of its neighbors. Under the local processing assumption, a cyclic structural dependency translates into mutual (causal) dependencies, \ie a potentially infinite loop, when computing the node states in parallel. The way to solve this is to assume an \textit{iterative} scheme, \ie the state $\mathbf{h}_v^{\ell+1}$ of node $v$ at iteration $\ell+1$ is defined using the neighbor states computed at the previous iteration $\ell$. The iterative scheme can be interpreted as a process that incrementally refines node representation as $\ell$ increases. While this might seem reasonable, one may question whether such an iterative process can converge, given the mutual dependencies among node states.
In practice, some approaches introduce constraints on the nature of the iterative process that force it to be convergent. Instead, others map each step of the iterative process to independent layers of a deep architecture. In other words, in the latter approach, the state $\mathbf{h}_v^{\ell+1}$ is computed by layer $\ell+1$ of the model based on the output of the previous layer $\ell$.

For the above reasons, in the following sections, we will use the symbol $\ell$ to refer, interchangeably, to an \textit{iteration step} or \textit{layer} by which nodes propagate information across the graph. Furthermore, we will denote with $\mathbf{h}_{g}^{\ell}$ the representation of the entire graph $g$ at layer $\ell$. 

\subsection{Three Mechanisms of Context Diffusion}
\label{sec:context-diffusion-mechanisms}

Another aspect of the process we have just discussed is the spreading of local information across the graph under the form of node states. This is arguably the most important concept of local and iterative graph learning methods. At a particular iteration $\ell$, we (informally) define the \textit{context} of a node state $\mathbf{h}_v^\ell$ as the set of node states that \textit{directly} or \textit{indirectly} contribute to determining $\mathbf{h}_v^\ell$; a formal characterization of context is given in \cite{micheli_neural_2009} for the interested reader.

An often employed formalism to explain how information is actually diffused across the graph is \textit{message passing} \cite{gilmer_neural_2017}. Focusing on a single node, message passing consists of two distinct operations:
\begin{itemize}
    \item \textit{message dispatching}. A message is computed for each node, using its current state and (possibly) edge information. Then, the message is sent to neighboring nodes according to the graph structure;
    \item \textit{state update}. The incoming node messages, and possibly its state, are collected and used to update the node state.
\end{itemize}
To bootstrap the message passing process, node states need to be initialized properly. A common choice is to set the initial states to their respective node feature vectors, although variations are possible. As discussed in Section \ref{sec:big-picture}, the order in which nodes are visited by the model to compute the states is not influential. The iterative application of message passing to the nodes allows contextual information to \quotes{travel} across the graph in the form of aggregated messages. As a consequence, nodes acquire knowledge about their wider surroundings rather than being restricted to their immediate neighborhood. \\

\begin{figure}[t]
\centering
\resizebox{\textwidth}{!}{\includegraphics{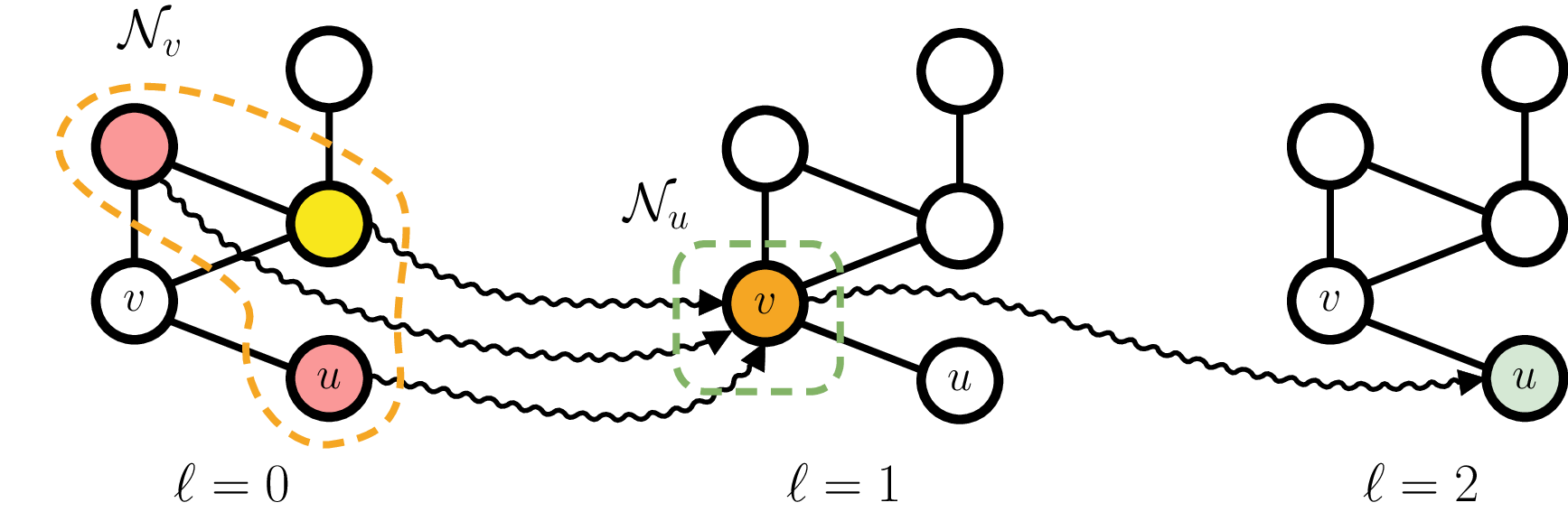}}
\caption{Context spreading in an undirected graph is shown for a network of depth 3, where wavy arrows represent the context flow. Specifically, we focus on the context of node $u$ at the last layer, by looking at the figure from right to left. It is easy to see that the context of node $u$ at $\ell=2$ depends on the state of its only neighbor $v$ at $\ell=1$, which in turn depends on its neighboring node representations at $\ell=0$ ($u$ included). Therefore, the context of $u$ is given by almost all the nodes in the graph.}\label{fig:context}
\end{figure}

Context diffusion can be visually represented in Figure \ref{fig:context}, in which we show the \quotes{view} that node $u$ has about the graph at iteration $\ell=2$. First of all, we observe that the neighborhood of $u$ is given by node $v$, and therefore all the contextual information that $u$ receives must go through $v$. If we look at the picture from right to left, $\mathbf{h}_u^2$ is defined in terms of $\mathbf{h}_v^1, \ v \in \mathcal{N}_u$,  which in turn is computed by aggregating three different colored nodes (one of which is $u$ itself). Hence, by iteratively computing local aggregation of neighbor states, we can indirectly provide $u$ with information about nodes farther away in the graph. It is trivial to show that $\ell=3$ iterations are sufficient to increase the context to include information from all nodes in the graph of Figure \ref{fig:context}. Put differently, not only deep learning techniques are useful for automatic feature extraction, but they are also \textit{functional to context diffusion}. \\

Under the light of the different information diffusion mechanisms they employ, we can partition most deep graph learning models into \textit{recurrent}, \textit{feedforward} and \textit{constructive} approaches. We now discuss how they work and what their differences are.

\paragraph{Recurrent Architectures} This family of models implements the iterative processing of node information as a dynamical system. Two of the most popular representatives of this family are the Graph Neural Network \cite{scarselli_graph_2009} and the Graph Echo State Network \cite{gallicchio_graph_2010}. Both approaches rely on imposing contractive dynamics to ensure convergence of the iterative process. While the former enforces such constraints in the (supervised) loss function, the latter inherits convergence from the contractivity of (untrained) reservoir dynamics. The Gated Graph Neural Network \cite{li_gated_2016} is another example of recurrent architecture where, differently from \cite{scarselli_graph_2009}, the number of iterations is fixed a priori, regardless of whether convergence is reached or not. An iterative approach based on \textit{collective inference}, which does not rely on any particular convergence criteria, was introduced in \cite{macskassy_classification_2007}. \\ This family of models handles graph cycles by modeling the mutual dependencies between node states using a single layer of recurrent units. In this case, we can interpret the symbol $\ell$ of Figure \ref{fig:context} as an \quotes{iteration step} of the recurrent state transition function computed for the state of each node. Finally, we mention the recent Fast and Deep Graph Neural Network \citep{gallicchio_fast_2020}, a multi-layered and efficient version of the Graph Echo State Network.

\paragraph{Feedforward Architectures}
In contrast to recurrent models, feedforward models do not exploit an iterative diffusion mechanism over the same layer of recurrent units. Instead, they stack multiple layers to \textit{compose} the local context learned at each step. As a result, the mutual dependencies induced by cycles are managed 
via differently parameterized layers without the need for constraints to ensure the convergence of the encoding process. To draw a parallel with Figure \ref{fig:context} (here $\ell$ corresponds to the index of a layer), this compositionality affects the context of each node, which increases as a function of the network depth up to the inclusion of the entire graph \cite{micheli_neural_2009}. The Neural Network for Graphs \cite{micheli_neural_2009} was the first proposal of a feedforward architecture for graphs. \\
Not surprisingly, there is a close similarity between this kind of context diffusion and the local receptive field of convolutional networks, which increases as more layers are added to the architecture. Despite that, the main difference is that graphs have no fixed structure as neighborhoods can vary in size, and a node ordering is rarely given. In particular, the local receptive field of convolutional networks can be seen as the context of a node in graph data, whereas the convolution operator processing corresponds to the  visit of the nodes in a graph (even though the parametrization technique is different). These are the reasons why the term \textit{graph convolutional layer} is often used in literature. \\ 
The family of feedforward models is the most popular for its simplicity, efficiency, and performance on many different tasks. However, deep networks for graphs suffer from the same gradient-related problems as other deep neural networks, especially when associated with an \quotes{end-to-end} learning process running through the whole architecture \citep{hochreiter_untersuchungen_1991, bengio_learning_1994, li_deeper_2018}. 

\paragraph{Constructive Architectures} The last family we identify can be seen as a special case of feedforward models, in which training is performed layer-wise. The major benefit of constructive architectures is that deep networks do not incur the vanishing/exploding gradient problem by design. Thus, the context can be more effectively propagated across layers and hence across node states. In supervised contexts, the constructive technique allows us to automatically determine the number of layers needed to solve a task \cite{fahlman_cascade-correlation_1990,marquez_deep_2018}. As explained in Figure \ref{fig:context}, this characteristic is also related to the context needed by the problem at hand; as such, there is no need to determine it {\em a priori}, as shown in \cite{micheli_neural_2009}, where the relationship between the depth of the layers and context shape is formally proved.

Moreover, an essential feature of constructive models is that they solve a problem in a \textit{divide-et-impera} fashion, incrementally splitting the task into more manageable sub-tasks (thus relaxing the \quotes{end-to-end} approach). Each layer contributes to the solution of a sub-problem, and subsequent layers use this result to solve the global task progressively.\\ 
Among the constructive approaches, we mention the Neural Network for Graphs \cite{micheli_neural_2009} (which is also the very first proposed feedforward architecture for graphs) and the Contextual Graph Markov Model \cite{bacciu_contextual_2018}, a more recent and probabilistic variant. 
\section{Building Blocks}
\label{sec:building-blocks}
We now turn our attention to the main constituents of local graph learning models. The architectural bias imposed by these building blocks determines the kind of representations that a model can compute. We remark that the aim of this Section is not to give the most comprehensive and general formulation under which all models can be formalized. Rather, it is intended to show the main \quotes{ingredients} that are common to many architectures and how these can be combined to compose an effective learning model for graphs. 

\subsection{Neighborhood Aggregation}
\label{sec:neighborhood-aggregation}
The way models aggregate neighbors to compute hidden node representations is at the core of local graph processing. We will conform to the common assumption that graphs are non-positional so that we need permutation invariant functions to realize the aggregation. For ease of notation, we will assume that any function operating on node $v$ has access to its feature vector $\mathbf{x}_v$, as well as the set of incident arc feature vectors, $\{\mathbf{a}_{uv} \mid u \in \mathcal{N}_v\}$.

In its most general form, neighborhood aggregation for node $v$ at layer/step $\ell+1$ can be represented as follows:
\begin{align}
\mathbf{h}_v^{\ell+1} = \phi^{\ell+1} \Big(\mathbf{h}_v^\ell,\ \Psi(\{\psi^{\ell+1}(\mathbf{h}_u^{\ell}) \mid u \in \mathcal{N}_v\} ) \Big)
\label{eq:simple-aggregation}
\end{align}
where $\mathbf{h}_u^{\ell}$ denotes the state of a node $u$ at layer/step $\ell$, $\phi$ and $\psi$ implement arbitrary transformations of the input data, \eg through a Multi Layer Perceptron, $\Psi$ is a permutation invariant function, and $\mathcal{N}_v$ can be the open or closed neighborhood of $v$. In most cases, the base case of $\ell=0$ corresponds to a possibly non-linear transformation of node features $\mathbf{x}_v$ which does not depend on structural information. \\ It is important to realize that the above formulation includes both Neural and Bayesian DGNs. As an example, a popular concrete instance of the neighborhood aggregation scheme presented above is the Graph Convolutional Network \cite{kipf_semi-supervised_2017}, a DNGN which performs aggregation as follows:
\begin{align}
& \mathbf{h}^{\ell+1}_v = \sigma(\mathbf{W}^{\ell+1} \sum_{u \in \mathcal{N}(v)}\mathbf{L}_{uv}\mathbf{h}^\ell_u),
\label{eq:gcn}
\end{align}
where $\mathbf{L}$ is the normalized graph Laplacian, $\mathbf{W}$ is a weight matrix and $\sigma$ is a non-linear activation function such as the sigmoid. We can readily see how Equation \ref{eq:gcn} is indeed a special case of Equation \ref{eq:simple-aggregation}:
\begin{align}
    & m_u^v = \psi^{\ell+1}(\mathbf{h}_u^{\ell}) \ \ \ \ \ \ \ \ \ \ \ = \mathbf{L}_{uv}\mathbf{h}^\ell_u \\
    & M_v = \Psi(\{m_u^v \mid u \in \mathcal{N}_v\}) = \sum_{u \in \mathcal{N}(v)}m_u^v \\
    & \mathbf{h}_v^{\ell+1} = \phi^{\ell+1} \Big(\mathbf{h}_v^\ell,\ M_v \Big) \ \ \ = \sigma(\mathbf{W}^{\ell+1} M_v).
\end{align}

In Section \ref{subsec:spectral-methods}, we will describe how the neighborhood aggregation of the Graph Convolutional Network is obtained via special approximations of spectral graph theory methodologies.

\paragraph{Handling Graph Edges}
The general neighborhood aggregation scheme presented above entails that arcs are unattributed or contain the same information. This assumption does not hold in general, as arcs in a graph often contain additional information about the nature of the relation. This information can be either discrete (\eg the type of chemical bonds that connect two atoms in a molecule) or continuous (\eg node distances between atoms). Thus, we need mechanisms that leverage arc labels to enrich node representations. If $\mathcal{A}$ is finite and discrete, we can reformulate Eq. \ref{eq:simple-aggregation} to account for different arc labels as follows:
\begin{align}
\mathbf{h}_v^{\ell+1} = \phi^{\ell+1} \Big(\mathbf{h}_v^\ell,\ \sum_{c_k \in \mathcal{A}}  \big( \Psi(\{\psi^{\ell+1}(\mathbf{h}_u^{\ell}) \mid u \in \mathcal{N}^{c_k}_v\} ) *w_{c_k} \big) \Big),
\label{eq:edge-aggregation-discrete}
\end{align}
where 
$w_{c_k}$ is a learnable scalar parameter that weighs the contribution of arcs with label $\mathbf{a}_{uv} = c_k$, and $*$ multiplies every component of its first argument by $w_{c_k}$. This formulation presents an inner aggregation among neighbors sharing the same arc label, plus an outer weighted sum over each possible arc label. This way, the contribution of each arc label is learned separately. The Neural Network for Graphs \cite{micheli_neural_2009} and the Relational Graph Convolutional Network \cite{schlichtkrull_modeling_2018} implement Eq. \ref{eq:edge-aggregation-discrete} explicitly, whereas the Contextual Graph Markov Model \cite{bacciu_contextual_2018} uses the switching-parent approximation \cite{saul_mixed_1999} to achieve the same goal. A more general solution, which works with continuous arc labels, is to reformulate Eq. \ref{eq:simple-aggregation} as
\begin{align}
\mathbf{h}_v^{\ell+1} = \phi^{\ell+1} \Big(\mathbf{h}_v^\ell,\ \Psi(\{e^{\ell+1}(\mathbf{a}_{uv})^T \psi^{\ell+1}(\mathbf{h}_u^{\ell}) \mid u \in \mathcal{N}_v\} ) \Big),
\label{eq:edge-aggregation-continuous}
\end{align}
where $e$ can be any function. Note how we explicitly introduce a dependence on the arc $\mathbf{a}_{uv}$ \textit{inside} the neighborhood aggregation: this has the effect of weighting the contribution of each neighbor based on its (possibly multidimensional) arc label, regardless of whether it is continuous or discrete. For example, in \cite{gilmer_neural_2017} $e$ is implemented as a neural network that outputs a weight matrix.

\paragraph{Attention}
Attention mechanisms \cite{vaswani_attention_2017} assign a relevance score to each part of the input of a neural layer, and they have gained popularity in language-related tasks. When the input is graph-structured, we can apply attention to the aggregation function. This results in a weighted average of the neighbors where individual weights are a function of node $v$ and its neighbor $u \in \mathcal{N}_v$. More formally, we extend the convolution of Eq. \ref{eq:simple-aggregation} in the following way:
\begin{align}
\mathbf{h}_v^{\ell+1} = \phi^{\ell+1} \Big(\mathbf{h}_v^\ell,\ \Psi(\{\alpha^{\ell+1}_{uv}*\psi^{\ell+1}(\mathbf{h}_u^{\ell}) \mid u \in \mathcal{N}_v\} ) \Big),
\label{eq:attention-aggregation}
\end{align}
where $\alpha^{\ell+1}_{uv} \in \mathbb{R}$ is the \emph{attention score} associated with $u \in \mathcal{N}_v$. In general, this score is unrelated to the edge information, and as such edge processing and attention are two quite distinct techniques. As a matter of fact, the Graph Attention Network \cite{velickovic_graph_2018} applies attention to its neighbors but it does not take into account edge information. To calculate the attention scores, the model computes \emph{attention coefficients} $w_{uv}$ as follows:
\begin{align}
    w_{uv}^{\ell} = a(\mathbf{W}^{\ell}\,\mathbf{h}_u^{\ell}, \mathbf{W}^{\ell}\,\mathbf{h}_v^{\ell}),
\end{align}
where $a$ is a shared attention function and $\mathbf{W}$ are the layer weights. The attention coefficients measure some form of similarity between the current node $v$ and each of its neighbors $u$. Moreover, the attention function $a$ is implemented as: 
\begin{align}
    a(\mathbf{W}^{\ell}\,\mathbf{h}_u^{\ell}, \mathbf{W}^{\ell}\,\mathbf{h}_v^{\ell}) = \mathrm{LeakyReLU}((\mathbf{b}^{\ell})^T [\mathbf{W}^{\ell}\,\mathbf{h}_u^{\ell}, \mathbf{W}^{\ell}\,\mathbf{h}_v^{\ell}]),
\end{align}
where $\mathbf{b}^{\ell}$ is a learnable parameter, $[\cdot,\cdot]$ denotes concatenation, and LeakyReLU is the non-linear activation function proposed in \cite{maas_rectifier_2013}. From the attention coefficients, one can obtain attention scores by passing them through a softmax function:
\begin{align}
    \alpha_{uv}^{\ell} = \frac{\exp(w^{\ell}_{uv})}{\sum_{u' \in \mathcal{N}_v} \exp(w^{\ell}_{u'v})}.
\end{align}
The Graph Attention Network also proposes a \textit{multi-head attention} technique, in which the results of multiple attention mechanisms are either concatenated or averaged together.

\begin{figure}[t]
\centering
\resizebox{\textwidth}{!}{\includegraphics{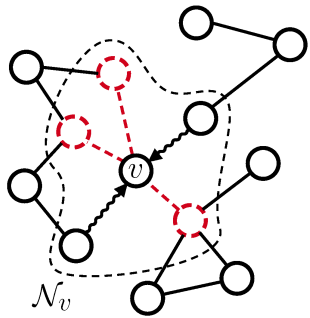}}
\caption{The sampling technique affects the neighborhood aggregation procedure by selecting either a subset of the neighbors \cite{chen_fastgcn_2018} or a subset of the nodes in the graph \cite{hamilton_inductive_2017} to compute $h_v^{\ell+1}$. Here, nodes in red have been randomly excluded from the neighborhood aggregation of node $v$, and the context flows only through the wavy arrows. }\label{fig:sampling}
\end{figure}

\paragraph{Sampling}
When graphs are large and dense, it can be unfeasible to perform aggregations over all neighbors for each node, as the number of edges becomes quadratic in $|\mathcal{V}_g|$. Therefore, alternative strategies are needed to reduce the computational burden, and neighborhood sampling is one of them. In this scenario, only a random subset of neighbors is used to compute $\mathbf{h}_v^{\ell+1}$. When the subset size is fixed, we also get an upper bound on the aggregation cost per graph. Figure \ref{fig:sampling} depicts how a generic sampling strategy acts at node level. Among the models that sample neighbors we mention Fast Graph Convolutional Network (FastGCN) \cite{chen_fastgcn_2018} and Graph SAmple and aggreGatE (GraphSAGE) \cite{hamilton_inductive_2017}. Specifically, FastGCN samples $t$ nodes at each layer $\ell$ via importance sampling so that the variance of the gradient estimator is reduced. Differently from FastGCN, GraphSAGE considers a neighborhood function $\mathcal{N}: |\mathcal{V}_g| \rightarrow 2^{|\mathcal{V}_g|}$ that associates each node with any (fixed) subset of the nodes in the given graph. In practice, GraphSAGE can sample nodes at multiple distances and treat them as direct neighbors of node $v$. Therefore, rather than learning locally, this technique exploits a wider and heterogeneous neighborhood, trading a potential improvement in performances for additional (but bounded) computational costs.

\subsection{Pooling}
\label{subsec:pooling}
\begin{figure}[t]
\centering
\resizebox{\textwidth}{!}{\includegraphics{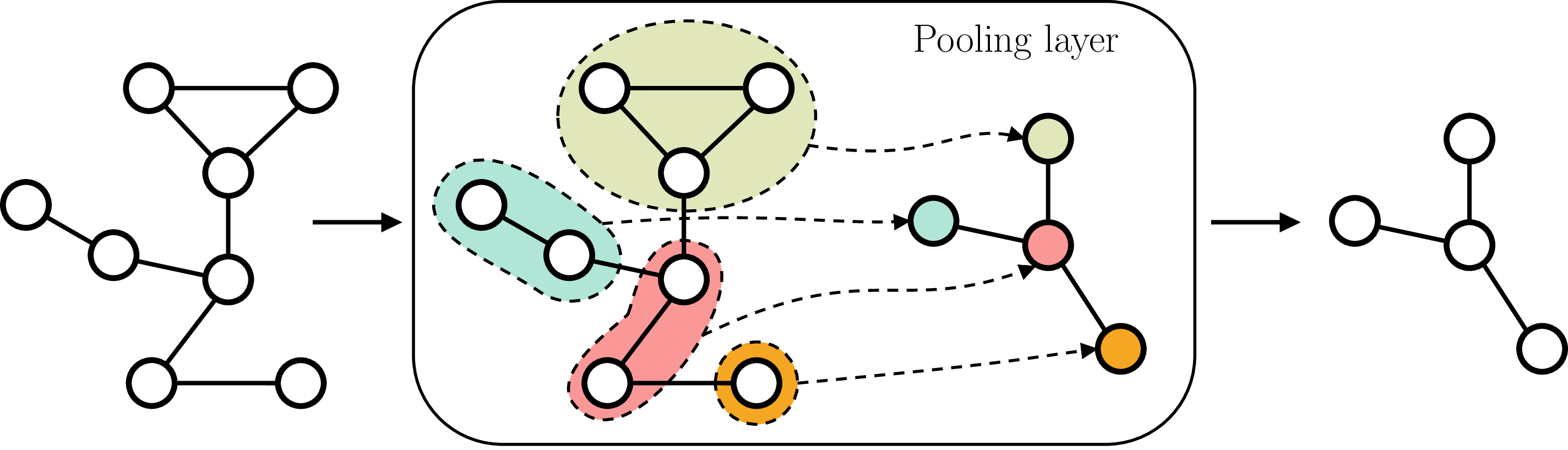}}
\caption{We show an example of the pooling technique. Each pooling layer coarsens the graph by identifying and clustering nodes of the same community together, so that each group becomes a node of the coarsened graph.}\label{fig:pooling}
\end{figure}

Similarly to convolutional networks for images, graph pooling operators can be defined to reduce the dimension of the graph after a DGN layer. Graph pooling is mainly used for three purposes, that is to discover important communities in the graph, to imbue this knowledge in the learned representations, and to reduce the computational costs in large scale structures. Figure \ref{fig:pooling} sketches the general idea associated with this techinque. Pooling mechanisms can be differentiated in two broad classes: \textit{adaptive} and \textit{topological}. The former relies on a parametric, and hence trainable, pooling mechanism. A notable example of this approach is Differentiable Pooling \cite{ying_hierarchical_2018}, which uses a neural layer to learn a clustering of the current nodes based on their embeddings at the previous layer. Such clustering is realized by means of a DNGN layer, followed by a softmax to obtain a soft-membership matrix $\mathbf{S}^{\ell + 1}$ that associates nodes with clusters:
\begin{align}
    \mathbf{S}^{\ell+1} = \mathrm{softmax}(\mathrm{GNN}(\mathbf{A}^{\ell}, \mathbf{H}^{\ell})),
\end{align}
where $\mathbf{A}^{\ell}$ and $\mathbf{H}^{\ell}$ are the adjacency and encoding matrices of layer $\ell$. The $\mathbf{S}^{\ell+1}$ matrix is then used to recombine the current graph into (ideally) one of reduced size:
\begin{align}\label{eq:diff-pool}
    \mathbf{H}^{\ell+1} = {\mathbf{S}^{\ell+1}}^T \mathbf{H}^{\ell} \quad\quad\text{and}\quad\quad  \mathbf{A}^{\ell+1} = {\mathbf{S}^{\ell+1}}^T \mathbf{A}^{\ell} {\mathbf{S}^{\ell+1}}.
\end{align}
In practice, since the cluster assignment is soft to preserve differentiability,  its application produces dense adjacency matrices. Top-k Pooling \cite{gao_graph_2019} overcomes this limitation by learning a projection vector $p^\ell$ that is used to compute projection scores of the node embedding matrix using dot product, \ie
\begin{equation}
    s^{\ell+1} = \frac{\mathbf{H}^{\ell}p^{\ell+1}}{\|p^{\ell+1}\|}.
\end{equation}
Such scores are then used to select the indices of the top ranking nodes and to slice the matrix of the original graph to retain only the entries corresponding to top nodes. Node selection is made differentiable by means of a gating mechanism built on the projection scores. Self-attention Graph Pooling \cite{lee_self-attention_2019} extends Top-k Pooling by computing the score vector as an attention score with a Graph Convolutional Network \cite{kipf_semi-supervised_2017} 
\begin{equation}
    s^{\ell+1} = \sigma(\mathrm{GCN}(\mathbf{A}^{\ell},\mathbf{H}^{\ell})).
\end{equation}
Edge Pooling \cite{frederik_diehl_towards_2019} operates from a different perspective, by targeting edges in place of nodes. Edges are ranked based on a parametric scoring function which takes in input the concatenated embeddings of the incident nodes, that is
\begin{equation}
    s^{\ell+1}((v, u) \in \mathcal{E}_g) = \sigma(\mathbf{w}^T [\mathbf{h}^{\ell}_v, \mathbf{h}^{\ell}_u] + \mathbf{b}).
\end{equation}
The highest ranking edge and its incident nodes are then contracted into a single new node with appropriate connectivity, and the process is iterated.

Topological pooling, on the other hand, is non-adaptive and typically leverages the structure of the graph itself as well as its communities. Note that, them being non-adaptive, such mechanisms are not required to be differentiable, and their results are not task-dependent. Hence, these methods are potentially reusable in multi-task scenarios. The graph clustering software (GRACLUS) \cite{dhillon_weighted_2007} is a widely used graph partitioning algorithm that leverages an efficient approach to spectral clustering. Interestingly, GRACLUS does not require an eigendecomposition of the adjacency matrix. From a similar perspective, Non-negative Matrix Factorization Pooling \cite{bacciu_non-negative_2019} provides a soft node clustering using a non-negative factorization of the adjacency matrix.

Pooling methods can also be used to perform graph classification by iteratively shrinking the graph up to the point in which the graph contains a single node. Generally speaking, however, pooling is interleaved with DGNs layers so that context can be diffused before the graph is shrunk.
 \subsection{Node Aggregation for Graph Embedding}
\label{sec:node-aggregation-graph-embedding}
If the task to be performed requires it, \eg graph classification, node representations can be aggregated in order to produce a global graph embedding. Again, since no assumption about the size of a given graph holds in general, the aggregation needs to be permutation-invariant. More formally, a graph embedding at layer $\ell$ can be computed as follows:
\begin{align}
    \mathbf{h}^{\ell}_g = \Psi \Big( \{ f(\mathbf{h}^{\ell}_v) \mid v \in \mathcal{V}_g \} \Big),
\end{align}
where a common setup is to take $f$ as the identity function and choose $\Psi$ among element-wise mean, sum or max. Another, more sophisticated, aggregation scheme draws from the work of \cite{zaheer_deep_2017}, where a family of adaptive permutation-invariant functions if defined. Specifically, it implements $f$ as a neural network applied to all the node representations in the graph, and $\Psi$ is an element-wise summation followed by a final non-linear transformation. 

There are multiple ways to exploit graph embeddings at different layers for the downstream tasks. A straightforward way is to use the graph embedding of the last layer as a representative for the whole graph. More often, all the intermediate embeddings are concatenated or given as input to permutation-invariant aggregators. The work of \cite{li_gated_2016} proposes a different strategy where all the intermediate representations are viewed as a sequence, and the model learns a final graph embedding as the output of a Long Short-Term Memory \cite{hochreiter_long_1997} network on the sequence. Sort Pooling \cite{zhang_end--end_2018}, on the other hand, uses the concatenation of the node embeddings of all layers as the continuous equivalent of node coloring algorithms. Then, such \quotes{colors} define a lexicographic ordering of nodes across graphs. The top ordered nodes are then selected and fed (as a sequence) to a one-dimensional convolutional layer that computes the aggregated graph encoding.   

\begin{figure}[ht]
\centering
\resizebox{0.95\textwidth}{!}{\includegraphics{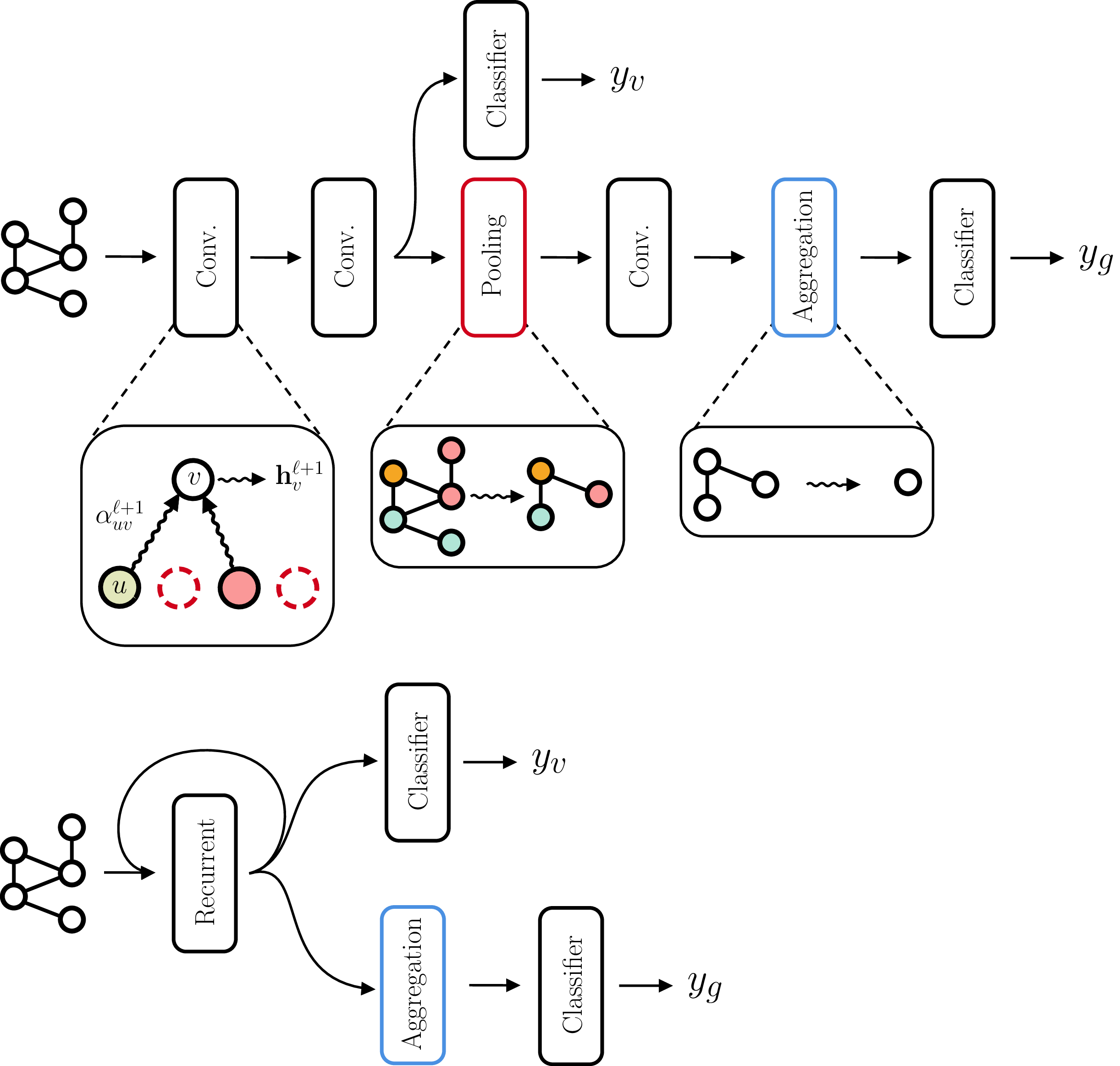}}
\caption{Two possible architectures (feedforward and recurrent) for node and graph classification. Inside each layer, one can apply the attention and sampling techniques described in this Section. After pooling is applied, it is not possible to perform node classification anymore, which is why a potential model for node classification can combine graph convolutional layers. A recurrent architecture (bottom) iteratively applies the same neighborhood aggregation, possibly until a convergence criterion is met.}\label{fig:example}
\end{figure}

To conclude, Table \ref{tab:summary-aggregation} provides a summary of neighborhood aggregation methods for some representative models.  Figure \ref{fig:example} visually exemplifies how the different building blocks can be arranged and combined to construct a feedforward or recurrent model that is end-to-end trainable.

\begin{table}[ht]
\centering
\small
\renewcommand{\arraystretch}{2.0}
\begin{tabular}{lc}
Model          & Neighborhood Aggregation $\mathbf{h}_v^{\ell+1}$  \\ \hline
NN4G \cite{micheli_neural_2009}      &   $\sigma \Big(\mathbf{w}^{\ell +1^T}\mathbf{x}_v + \sum_{i=0}^{\ell}\sum_{c_k \in \mathcal{C}} \sum_{u \in \mathcal{N}^{c_k}_v} w_{c_k}^i * \mathbf{h}_u^i \Big)$ \\
GNN \cite{scarselli_graph_2009}       &  $\sum_{u \in \mathcal{N}_v} MLP^{\ell+1}\Big(\mathbf{x}_u, \mathbf{x}_v, \mathbf{a}_{uv}, \mathbf{h}_u^\ell \Big) $  \\
GraphESN \cite{gallicchio_graph_2010}   &  $\sigma \Big(\mathbf{W}^{\ell+1}\mathbf{x}_u +\hat{\mathbf{W}}^{\ell+1}[\mathbf{h}^{\ell}_{u_1}, \dots, \mathbf{h}^{\ell}_{u_{\mathcal{N}_v}}]\Big) $  \\
GCN \cite{kipf_semi-supervised_2017}       &     $\sigma \Big(\mathbf{W}^{\ell +1} \sum_{u \in \mathcal{N}(v)} \mathbf{L}_{vu}\mathbf{h}^\ell_u \Big)$ \\
GAT \cite{velickovic_graph_2018}      &   $\sigma\Big(\sum_{u \in \mathcal{N}_v} \alpha_{uv}^{\ell+1}*\mathbf{W}^{\ell+1}\mathbf{h}_u\Big)$     \\
ECC \cite{simonovsky_dynamic_2017} &    $\sigma\Big( \frac{1}{|\mathcal{N}_v|}\sum_{u \in \mathcal{N}_v} MLP^{\ell+1}(\mathbf{a}_{uv})^T \mathbf{h}_u^\ell \Big) $   \\
R-GCN \cite{schlichtkrull_modeling_2018}       &    $\sigma \Big(\sum_{c_k \in \mathcal{C}}\sum_{u \in \mathcal{N}_v^{c_k}} \frac{1}{|\mathcal{N}_v^{c_k}|}\mathbf{W}_{c_k}^{\ell +1} \mathbf{h}_u^\ell + \mathbf{W}^{\ell +1} \mathbf{h}_v^\ell\Big)$  \\
GraphSAGE \cite{hamilton_inductive_2017} &  $\sigma\Big(\mathbf{W}^{\ell+1}(\frac{1}{|\mathcal{N}_v|}[\mathbf{h}_v^\ell, \sum_{u \in \mathcal{N}_v} \mathbf{h}_u^\ell])\Big)  $  \\
CGMM \cite{bacciu_contextual_2018}      &  $ \sum_{i=0}^{\ell} w^i * \Big(\sum_{c_k \in \mathcal{C}} w_{c_k}^i * \big( \frac{1}{|\mathcal{N}_v^{c_k}|}\sum_{u \in \mathcal{N}^{c_k}_v} \mathbf{h}_u^i \big) \Big) $ \\ 
GIN \cite{xu_how_2019}       & $MLP^{\ell+1} \Big( \big(1 + \epsilon^{\ell+1} \big)\mathbf{h}_v^{\ell} + \sum_{u \in \mathcal{N}_v} \mathbf{h}_u^\ell \Big)$   \\
\end{tabular}
\caption{We report some of the neighborhood aggregations present in the literature, and we provide a table in \ref{appendix:A} to ease referencing and understanding of acronyms. Here, square brackets denote concatenation, and $W, w$ and $\epsilon$ are learnable parameters. Note that GraphESN assumes a maximum size of the neighborhood. The attention mechanism of GAT is implemented by a weight $\alpha_{uv}$ that depends on the associated nodes. As for GraphSAGE, we describe its \quotes{mean} variant, though others have been proposed by the authors. Finally, recall that $\ell$ represents an \textit{iteration step} in GNN rather than a layer.}
\label{tab:summary-aggregation}
\end{table}

\section{Learning Criteria}
\label{sec:learning-criteria}
After having introduced the main building blocks and most common techniques to produce node and graph representations, we now discuss the different learning criteria that can be used and combined to tackle different tasks. We will focus on unsupervised, supervised, generative, and adversarial learning criteria to give a comprehensive overview of the research in this field.

\subsection{Unsupervised Learning}
Our discussion begins with unsupervised learning criteria, as some of them act as regularizers in more complex objective functions.

\paragraph{Link Prediction} The most common unsupervised criterion used by graph neural networks is the so-called \textit{link prediction} or \textit{reconstruction} loss. This learning objective aims at building node representations that are similar if an arc connects the associated nodes, and it is suitable for link prediction tasks. Formally, the reconstruction loss can be defined \cite{kipf_semi-supervised_2017} as
\begin{align}
    \mathcal{L}_{rec}(g) = \sum_{(u,v)}||\mathbf{h}_v - \mathbf{h}_u||^2.
\end{align}
There also exists a probabilistic formulation of this loss, which is used in variational auto-encoders for graphs \cite{kipf_variational_2016} where the decoder only focuses on structural reconstruction:
\begin{align}\label{eq:structural-loss-prob}
P( (u,v) \in \mathcal{E}_g \mid \mathbf{h}_u, \mathbf{h}_v) = \sigma(\mathbf{h}_u^T\mathbf{h}_v),
\end{align}
where $\sigma$ is the sigmoid function (with co-domain in $[0,1]$). \\ Importantly, the link prediction loss reflects the assumption that neighboring nodes should be associated to the same class/community, which is also called \textit{homphily} \cite{macskassy_classification_2007}. In this sense, this unsupervised loss can be seen as a regularizer to be combined with other supervised loss functions. In all tasks where the homophily assumption holds, we expect this loss function to be beneficial.

\paragraph{Maximum Likelihood} When the goal is to build unsupervised representations that reflect the \textit{distribution} of neighboring states, a different approach is needed. In this scenario, probabilistic models can be of help. Indeed, one can compute the likelihood that node $u$ has a certain label $\mathbf{x}_u$ conditioned on neighboring information. Known unsupervised probabilistic learning approaches can then maximize this likelihood. An example is the Contextual Graph Markov Model \cite{bacciu_contextual_2018}, which constructs a deep network as a stack of simple Bayesian networks. Each layer maximizes the following likelihood:
\begin{align}
\mathcal{L}(\theta|g) = \prod_{ u \in \mathcal{V}_g} \sum_{i=1}^{C} P(y_u | Q_u = i) P(Q_u = i | \mathbf{q}_{\mathcal{N}_u}),
\end{align}where $Q_u$ is the categorical latent variable with $C$ states associated to node $u$, and $\mathbf{q}_{\mathcal{N}_u}$ is the set of neighboring states computed so far. On the other hand, there are hybrid methods that maximize an intractable likelihood with a combination of variational approximations and DNGNs \cite{qu_gmnn_2019, kipf_variational_2016}.

Maximum likelihood can also be used in the more standard unsupervised task of density estimation. In particular, a combination of a graph encoder with a radial basis function network can be jointly optimized to solve both tasks \cite{trentin_maximum_2009, bongini_recursive_2018}. Interestingly, the formulation also extends the notion of random graphs \cite{gilbert_random_1959,erdhos_evolution_1960} to a broader class of graphs to define probability distributions on attributed graphs, and under some mild conditions it even possesses universal approximation capabilities.

\paragraph{Graph Clustering}
Graph clustering aims at partitioning a set of graphs into different groups that share some form of similarity. Usually, similarity can be achieved by a distance-based criterion working on vectors obtained via graph encoders. A large family of well-known approaches for directed acyclic graphs, most of which are based on Self-Organizing Maps \cite{kohonen_self_1990}, has been reviewed and studied in \cite{hagenbuchner_self_2003, hammer_recursive_2004, hammer_general_2004}. These foundational works were later extended to deal with more cyclic graphs \cite{neuhaus_self_2005, hagenbuchner_graph_2009} Finally, the maximum-likelihood based technique in \cite{bongini_recursive_2018} can be straightforwardly applied to graph clustering.

\paragraph{Mutual Information} An alternative approach to produce node representations focuses on local mutual information maximization between pairs of graphs. In particular, Deep Graph Infomax \cite{velickovic_deep_2019} uses a corruption function that generates a distorted version of a graph $g$, called $\tilde{g}$. Then, a discriminator is trained to distinguish the two graphs, using a bilinear score on node and graph representations. This unsupervised method requires a corruption function to be manually defined each time, \eg injecting random structural noise in the graph, and as such it imposes a bias on the learning process.

\paragraph{Entropy regularization for pooling} When using adaptive pooling methods, it can be useful to encourage the model to assign each node to a single community. Indeed, adaptive pooling can easily scatter the contribution of node $u$ across multiple communities, and this results in low informative communities. The \textit{entropy} loss was proposed \cite{ying_hierarchical_2018} to address this issue. Formally, if we define with $\mathbf{S} \in \mathbb{R}^{|\mathcal{V}_g|\times C}$ the matrix of soft-cluster assignments (Section \ref{subsec:pooling}, where $C$ is the number of clusters of the pooling layer, the entropy loss is computed as follows:
\begin{align}
\mathcal{L}_{ent}(g)=\frac{1}{|\mathcal{V}_g|}\sum_{u \in \mathcal{V}_g}H(\mathbf{S}_u)
\end{align}
where $H$ is the entropy and $\mathbf{S}_u$ is the row associated with node $u$ clusters assignment. Notice that, from a practical point of view, it is still challenging to devise a differentiable pooling method that does not generate dense representations. However, encouraging a one-hot community assignment of nodes can enhance visual interpretation of the learned clusters, and it acts as a regularizer that enforces well-separated communities.

\subsection{Supervised Learning}
We logically divide supervised graph learning tasks in node classification, graph classification, and graph regression. Once node or graph representations are learned, the prediction step does not differ from standard vectorial machine learning, and common learning criteria are Cross-Entropy/Negative Log-likelihood for classification and Mean Square Error for regression.

\paragraph{Node Classification}
As the term indicates, the goal of node classification is to assign the correct target label to each node in the graph. There can be two distinct settings: \textit{inductive node classification}, which consists of classifying nodes that belong to unseen graphs, and \textit{transductive node classification}, in which there is only one graph to learn from and only a fraction of the nodes needs to be classified. It is important to remark that benchmark results for node classification have been severely affected by delicate experimental settings; this issue was later addressed \cite{shchur_pitfalls_2018} by re-evaluating state of the art architectures under a rigorous setting. Assuming a multi-class node classification task with $C$ classes, the most common learning criterion is the cross-entropy:
\begin{align}
    \mathcal{L}_{CE}(y, t) = -\log\Big( \frac{e^{y_t}}{\sum_{j=1}^C e^{y_j}} \Big)
    \label{eq:cross-entropy}
\end{align}
where $y \in R^{C}$ and $t \in \{1,\dots,C\}$ are the output vector and target class, respectively. The loss is then summed or averaged over all nodes in the dataset.

\paragraph{Graph Classification/Regression}
To solve graph classification and regression tasks, it is first necessary to apply the node aggregation techniques discussed in Section \ref{sec:node-aggregation-graph-embedding}. After having obtained a single graph representation, it is straightforward to perform classification or regression via standard machine learning techniques. Similarly to node classification, the graph classification field suffers from ambiguous, irreproducible, and flawed experimental procedures that have been causing a great deal of confusion in the research community. Very recently, however, it has been proposed a rigorous re-evaluation of state-of-the-art models across a consistent number of datasets \cite{errica_fair_2020} aimed at counteracting this troubling trend. Cross entropy is usually employed for multi-class graph classification, whereas Mean Square Error is a common criterion for graph regression:
\begin{align}
    \mathcal{L}_{MSE}(y, t) = \frac{1}{|\mathcal{G}|}||y-t||_2^2
    \label{eq:mse}
\end{align}
where $\mathcal{G}$ represents the dataset the dataset and $y,t$ are the output and target vectors, respectively. Again, the loss is summed or averaged over all graphs in the dataset.

\subsection{Generative learning}
\begin{figure}[th]
\centering
\resizebox{\textwidth}{!}{\includegraphics{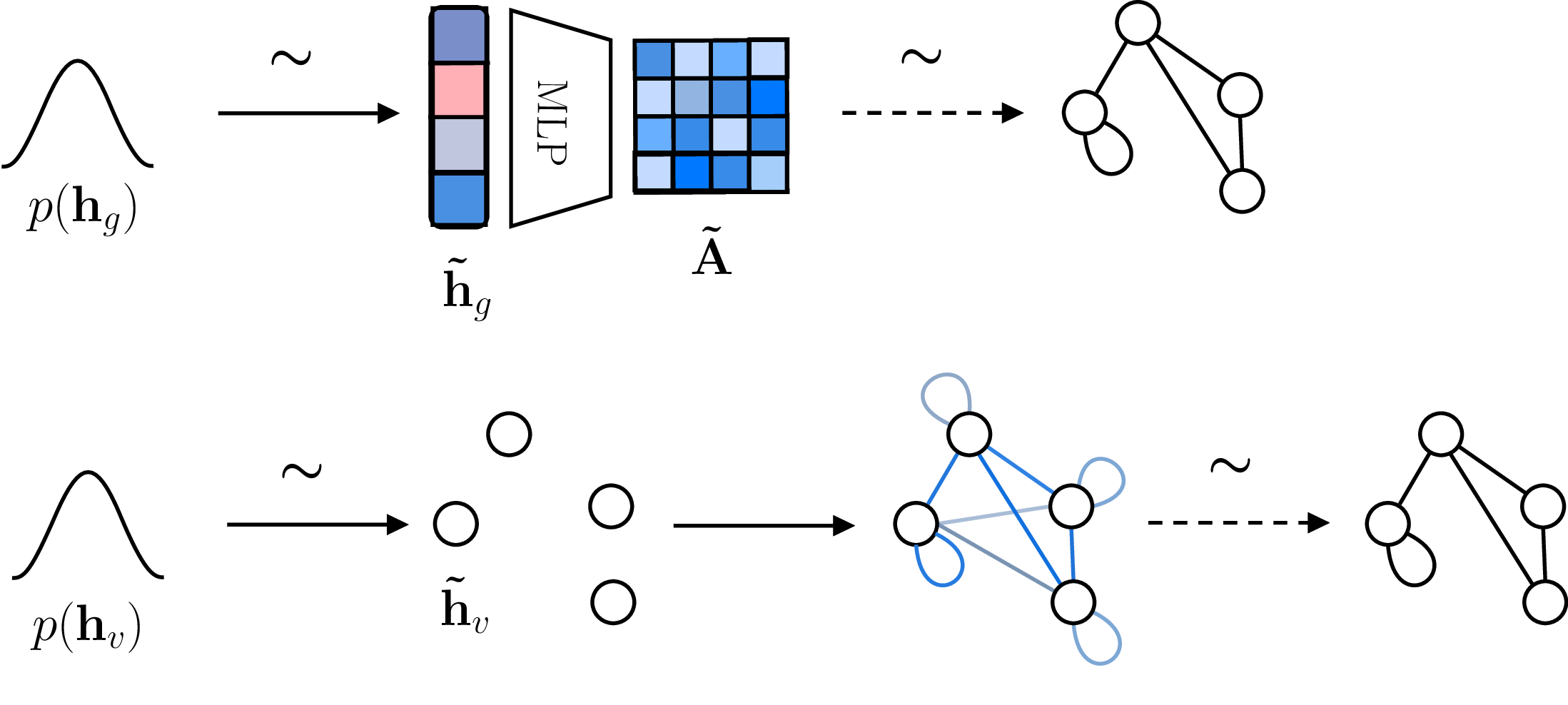}}
\caption{A simplified schema of graph-level (top row) and node-level (bottom row) generative decoders is shown. Tilde symbols on top of arrows indicate sampling. Dashed arrows indicate that the corresponding sampling procedure is not differentiable in general. Darker shades of blue indicate higher probabilities.}\label{fig:generative}
\end{figure}

Learning how to generate a graph from a dataset of available samples is arguably a more complex endeavor than the previous tasks. To sample a graph $g$, one must have access to the underlying generating distribution $P(g)$. However, since graph structures are discrete, combinatorial, and of variable-size, gradient-based approaches that learn the marginal probability of data are not trivially applicable. Thus, the generative process is conditioned on a latent representation of a graph/set of nodes, from which the actual structure is decoded. We now present the two most popular approaches by which DGGNs decode graphs latent samples, both of which are depicted in Figure \ref{fig:generative}. For ease of comprehension, we will focus on the case of graphs with unattributed nodes and edges. Crucially, we assume knowledge of a proper sampling technique; later on, we discuss how these sampling mechanisms can even be learned. 

\paragraph{Graph-level decoding} These approaches sample the graph adjacency matrix in one shot. More in detail, the decoder takes a graph representation as input, and it outputs a dense probabilistic adjacency matrix $\mathbf{\tilde{A}} \in \mathbb{R}^{k \times k}$, where $k$ is the maximum number of nodes allowed, and each entry $\tilde{a}_{ij}$ specifies the probability of observing an arc between node $i$ and $j$. This corresponds to minimizing the following log-likelihood:
\begin{align}
    \mathcal{L}_{\mathrm{decoder}}(g) = -\log P(\mathbf{\tilde{A}}\mid\mathbf{\tilde{h}}_g),
\end{align}
where $\mathbf{\tilde{h}}_g$ is a sampled graph representation and $P(\mathbf{\tilde{A}}\mid\mathbf{\tilde{h}}_g)$ is implemented as a multi-layer perceptron applied to $\mathbf{\tilde{h}}_g$. To obtain a novel graph, one can either:
\begin{enumerate}
    \item sample each entry of the probabilistic adjacency matrix, with connection probability $\tilde{a}_{ij}$;
    \item perform an approximate graph matching between the probabilistic and the ground truth matrices, as in \cite{simonovsky_graphvae_2018, kwon_efficient_2019};
    \item make the sampling procedure differentiable using a categorical reparameterization with a Gumbel-Softmax \cite{jang_categorical_2017}, as explored for example in \cite{de_cao_molgan_2018}.
\end{enumerate} 
Notice that the first two alternatives are not differentiable; in those cases, the actual reconstruction loss cannot be back-propagated during training. Thus, the reconstruction loss is computed on the probabilistic matrix instead of the actual matrix \cite{simonovsky_graphvae_2018}. Graph-level decoders are not permutation invariant (unless approximate graph matching is used) because the ordering of the output matrix is assumed fixed. 
\paragraph{Node-level decoding} Node-level decoders generate a graph starting from a set of $k$ node representations. These are sampled according to an approximation of their probability distribution. To decode a graph in this setting, one needs to generate the adjacency matrix conditioned on the sampled node set. This is achieved by introducing all possible $k(k+1)/2$ unordered node pairs as input to a decoder that optimizes the following log-likelihood:
\begin{align}\label{eq:node-level-decoder}
    \mathcal{L}_{\mathrm{decoder}}(g) = -\frac{1}{\vert\mathcal{V}_g\vert} \sum_{v \in \mathcal{V}_g} \sum_{u \in \mathcal{V}_g}\log P(\tilde{a}_{uv} \mid \mathbf{\tilde{h}}_v, \mathbf{\tilde{h}}_u),
\end{align}
where $P(\tilde{a}_{uv} \mid \mathbf{\tilde{h}}_v, \mathbf{\tilde{h}}_u) = \sigma(\mathbf{\tilde{h}}_v^T\ \mathbf{\tilde{h}}_u)$ as in Eq. \ref{eq:structural-loss-prob} and similarly to \cite{kipf_variational_2016, grover_graphite_2019}, and $\mathbf{\tilde{h}}$ are sampled node representations. As opposed to graph-level decoding, this method is permutation invariant, even though it is generally more expensive to calculate than one-shot adjacency matrix generation.
\\\\
To complement our discussion, in the following, we summarize those generative models that can optimize the decoding objective while jointly learning how to sample the space of latent representations. We distinguish approaches that explicitly learn their (possibly approximated) probability distribution from those that implicitly learn how to sample from the distribution. The former are based on Generative Auto-Encoders \cite{kingma_auto-encoding_2014, tolstikhin_wasserstein_2018}, while the latter leverage Generative Adversarial Networks \cite{goodfellow_generative_2014}.

\paragraph{Generative Auto-Encoder for graphs} This method works by learning the probability distribution of node (or graph) representations in latent space. Samples of this distribution are then given to the decoder to generate novel graphs. A general formulation of the loss function for graphs is the following:
\begin{align}
    \mathcal{L}_{\mathrm{AE}}(g) = \mathcal{L}_{\mathrm{decoder}}(g) + \mathcal{L}_{\mathrm{encoder}}(g),
    \label{eq:gae-loss}
\end{align}
where $\mathcal{L}_{\mathrm{decoder}}$ is the reconstruction error of the decoder as mentioned above, and $\mathcal{L}_{\mathrm{encoder}}$ is a divergence measure that forces the distribution of points in latent space to resemble a \quotes{tractable} prior (usually an isotropic Gaussian $\mathcal{N}(\mathbf{0}, \mathbf{I})$). For example, models based on Variational AEs \cite{kingma_auto-encoding_2014} use the following encoder loss:
\begin{align}
\mathcal{L}_{\mathrm{encoder}}(g) = -D_{KL}[\mathcal{N}(\boldsymbol{\mu}, \boldsymbol{\sigma}^2)\, \|\, \mathcal{N}(\mathbf{0}, \mathbf{I})],
\end{align}
where $D_{KL}$ is the Kullback-Leibler divergence, and the two parameters of the encoding distribution are computed as $\boldsymbol{\mu} = \mathrm{DGN}_{\mu}(\mathbf{A}, \mathbf{X})$ and $\boldsymbol{\sigma} = \mathrm{DGN}_{\sigma}(\mathbf{A}, \mathbf{X})$ \cite{simonovsky_graphvae_2018,liu_constrained_2018, samanta_nevae_2019}. More recent approaches such as \cite{bradshaw_model_2019} propose to replace the encoder error term in Equation \ref{eq:gae-loss} with a Wasserstein distance term \cite{tolstikhin_wasserstein_2018}.

\paragraph{Generative Adversarial Networks for graphs} This technique is particularly convenient. It does not work with $P(g)$ directly, but it only learns an adaptive mechanism to sample from it. Generally speaking, Generative Adversarial Networks use two different functions: a \emph{generator} $G$, which generates novel graphs, and a \emph{discriminator} $D$ that is trained to recognize whether its input comes from the generator or from the dataset. When dealing with graph-structured data, both the generator and the discriminator are trained jointly to minimize the following objective:
\begin{align}
    \mathcal{L}_{\mathrm{GAN}}(g) = \min_{G} \max_{D} \mathbb{E}_{g \sim P_{\mathrm{data}}(g)}[\log D(g)] + \mathbb{E}_{\mathbf{z} \sim P(\mathbf{z})} [\log(1- D(G(\mathbf{z})))],
\end{align}
where $P_{\mathrm{data}}$ is the true unknown probability distribution of the data, and $P(\mathbf{z})$ is the prior on the latent space (usually isotropic Gaussian or uniform). Note that this procedure provides an implicit way to sample from the probability distribution of interest without manipulating it directly. In the case of graph generation, $G$ can be a graph or node-level decoder that takes a random point in latent space as input and generates a graph, while $D$ takes a graph as input and outputs the probability of being a \quotes{fake} graph produced by the generator. As an example, \cite{fan_conditional_2019} implements $G$ as a graph-level decoder that outputs both a probabilistic adjacency matrix $\tilde{\mathbf{A}}$ and a node label matrix $\tilde{\mathbf{L}}$ as well. The discriminator takes an adjacency matrix $\mathbf{A}$ and a node label matrix $\mathbf{L}$ as input, applies a Jumping Knowledge Network \cite{xu_representation_2018} to it, and decides whether the graph is sampled from the generator or the dataset with a multi-layer perceptron. In contrast, \cite{wang_graphgan_2018} works at the node level. Specifically, $G$ generates structure-aware node representations (based on the connectivity of a breadth-first search tree of a random graph sampled from the training set), while the discriminator takes as input two node representations and decides whether they come from the training set or the generator, optimizing an objective function similar to Eq. \ref{eq:node-level-decoder}.  
\subsection{Summary}
\begin{table}[ht!]
\footnotesize
\centering
\begin{tabular}{lcccc}
Model & Context & Embedding & Layers   &  Nature  \\ \hline
GNN \cite{scarselli_graph_2009}     & Recurrent & Supervised & Single & Neural  \\
NN4G \cite{micheli_neural_2009}     & Constructive   & Supervised  & Adaptive & Neural \\
GraphESN \cite{gallicchio_graph_2010} & Recurrent & Untrained & Single & Neural  \\
GCN \cite{kipf_semi-supervised_2017}     & Feedforward    &    Supervised          &    Fixed        &  Neural   \\
GG-NN \cite{li_gated_2016} & Recurrent    &    Supervised          &    Fixed        &  Neural   \\
ECC \cite{simonovsky_dynamic_2017}     &        Feedforward         &  Supervised    &  Fixed & Neural \\
GraphSAGE \cite{hamilton_inductive_2017}     &        Feedforward         &  Both    &      Fixed & Neural \\
CGMM \cite{bacciu_contextual_2018}     &       Constructive         &  Unsupervised    &      Fixed & Probabilistic \\
DGCNN \cite{zhang_end--end_2018}    &        Feedforward     &  Supervised    &      Fixed & Neural \\    
DiffPool \cite{ying_hierarchical_2018}     &        Feedforward         &  Supervised    &  Fixed & Neural \\
GAT \cite{velickovic_graph_2018}    & Feedforward    &    Supervised          &       Fixed           & Neural          \\                     
R-GCN \cite{schlichtkrull_modeling_2018}    &       Feedforward         &  Supervised    &      Fixed & Neural \\
DGI \cite{velickovic_deep_2019}     &   Feedforward    &    Unsupervised          &       Fixed           & Neural \\
GMNN \cite{qu_gmnn_2019}     &   Feedforward    &    Both          &       Fixed           & Hybrid \\
GIN \cite{xu_how_2019}     &   Feedforward    &    Supervised          &       Fixed           & Neural \\
NMFPool \cite{bacciu_non-negative_2019}     &        Feedforward         &  Supervised    &  Fixed & Neural \\
SAGPool \cite{lee_self-attention_2019}     &        Feedforward         &  Supervised    &  Fixed & Neural \\
Top-k Pool \cite{gao_graph_2019}     &        Feedforward         &  Supervised    &  Fixed & Neural \\
FDGNN \cite{gallicchio_fast_2020}     &   Recurrent    &    Untrained      &       Fixed           & Neural \\ \\
\end{tabular}
\begin{tabular}{lcccc}
Model & Edges        & Pooling  &                    Attention                  & Sampling                   \\  \hline
GNN \cite{scarselli_graph_2009}    & Continuous    & \ding{55} & \ding{55} & \ding{55} \\
NN4G \cite{micheli_neural_2009}     &Discrete    & \ding{55} & \ding{55} & \ding{55} \\
GraphESN \cite{gallicchio_graph_2010}    & \ding{55} & \ding{55} & \ding{55} & \ding{55} \\
GCN \cite{kipf_semi-supervised_2017}     &   \ding{55}             &    \ding{55}    &            \ding{55}    &   \ding{55} \\
GG-NN \cite{li_gated_2016} &  \ding{55}             &    \ding{55}    &            \ding{55}    &   \ding{55} \\
ECC \cite{simonovsky_dynamic_2017}     &        Continuous         &  Topological    &      \ding{55} & \ding{55} \\
GraphSAGE \cite{hamilton_inductive_2017}    &        \ding{55}         &  \ding{55}    &      \ding{55} & \ding{51} \\
CGMM \cite{bacciu_contextual_2018}     &        Discrete       &  \ding{55}    &      \ding{55} & \ding{55} \\
DiffPool \cite{ying_hierarchical_2018}    &        -         & Adaptive   &     -    &    - \\
DGCNN \cite{zhang_end--end_2018}     &        \ding{55}         & Topological   &      \ding{55} & \ding{55} \\
GAT \cite{velickovic_graph_2018}     &        \ding{55}         &  \ding{55}    &      \ding{51} & \ding{55} \\
R-GCN \cite{schlichtkrull_modeling_2018}     &        Discrete         &  \ding{55}    &      \ding{55} & \ding{55} \\
GMNN \cite{qu_gmnn_2019} &       -         &  -    &     - & - \\
DGI \cite{velickovic_deep_2019}     &        \ding{55}         &  \ding{55}    &      \ding{55} & \ding{51} \\
GIN \cite{xu_how_2019}     &        \ding{55}         &  \ding{55}    &      \ding{55} & \ding{55} \\
NMFPool \cite{bacciu_non-negative_2019}    &        -         & Topological   &     -    &    - \\
SAGPool \cite{lee_self-attention_2019}    &        -         & Adaptive   &     -    &    - \\
Top-k Pool \cite{gao_graph_2019}    &        -         & Adaptive   &     -    &    - \\
FDGNN \cite{gallicchio_fast_2020}  &     \ding{55}    &     \ding{55}          &         \ding{55}           &   \ding{51} \\
\end{tabular}
\caption{Here we recap the main properties of DGNs, according to what we have discussed so far. Please refer to \ref{appendix:A} for a description of all acronyms. For clarity, \quotes{-} means not applicable, as the model is a framework that relies on any generic learning methodology. The \quotes{Layers} column describes how many layers are used by an architecture, which can be just one, a fixed number or adaptively determined by the learning process. On the other hand, \quotes{Context} refers to the context diffusion method of a specific layer, which was discussed in Section \ref{sec:context-diffusion-mechanisms}.}
\label{tab:overview-models}
\end{table}

We conclude this Section by providing a characterization of some of the local iterative models in accord with the building blocks and learning criteria discussed so far. Precisely, Table \ref{tab:overview-models} differentiates models with respect to four key properties, namely the context diffusion method, how an embedding is computed, how layers are constructed, and the nature of the approach. Then, we added other properties that a model may possess or not, such as the ability to handle edges, to perform pooling, to attend over neighbors, and to sample neighbors.

\section{Summary of Other Approaches and Tasks}\label{sec:otherworks}
There are several approaches and topics that are not covered by the taxonomy discussed in earlier sections. In particular, we focused our attention on deep learning methods for graphs, which are mostly based on local and iterative processing. For completeness of exposition, we now briefly review some of the topics that were kept out.

\subsection{Kernels} There is a long-standing and consolidated line of research related to kernel methods applied to graphs \cite{ralaivola_graph_2005, vishwanathan_graph_2010, shervashidze_weisfeiler-lehman_2011, frasconi_klog_2014, yanardag_deep_2015}. A kernel is informally defined as a generalized form of positive-definite function that computes similarity scores between pairs of inputs. A crucial aspect of kernel methods, which impacts their application to graphs, is that they are usually non-local and non-adaptive, \ie they require humans to design the kernel function. When applied to graphs, kernel methods work particularly well when the properties of interest are known, and it is still difficult to perform better with adaptive approaches. However, as mentioned above, non-adaptivity constitutes the main drawback of kernels, as it is not always clear which features we want to extract from the graph. Moreover, kernels suffer from scalability issues when the number of inputs in the dataset is too large (albeit with some exceptions, see \cite{shervashidze_weisfeiler-lehman_2011}). Importantly, kernel similarity matrices can be combined with Support Vector Machines \cite{cortes_support-vector_1995} to perform graph classification. Finally, we mention the Nonparametric Small Random Networks algorithm \citep{trentin_nonparametric_2018}, a recent technique for efficient graph classification. Despite being related to the 2-graphlet graph kernel \citep{shervashidze_efficient_2009}, its formulation is probabilistic, and it outperforms many DGNs and graph kernels on a number of tasks.

\subsection{Spectral methods}\label{subsec:spectral-methods} Spectral graph theory studies the properties of a graph by means of the associated adjacency and Laplacian matrices. Many machine learning problems can be tackled with these techniques, for example Laplacian smoothing \cite{sadhanala_graph_2016}, graph semi-supervised learning \cite{chapelle_semi-supervised_2006, calandriello_improved_2018} and spectral clustering \cite{von_luxburg_tutorial_2007}. A graph can also be analyzed with signal processing tools, such as the Graph Fourier Transform \cite{hammond_wavelets_2011} and related adaptive techniques \cite{bruna_spectral_2014}. Generally speaking, spectral techniques are meant to work on graphs with the same shape and different node labels, as they are based on the eigen-decomposition of adjacency and Laplacian matrices. More in detail, the eigenvector matrix $Q$ of the Laplacian constitutes an orthonormal basis used to compute the Graph Fourier Transform on the nodes signal $\mathbf{f} \in \mathbb{R}^{\mathcal{V}_g}$. The transform is defined as $\mathcal{F}(\mathbf{f}) = Q^T \mathbf{f}$, and its inverse is simply $\mathcal{F}^{-1}(Q^T\mathbf{f}) = QQ^T \mathbf{f}$ thanks to orthogonality of $Q$. Then, the graph convolution between a filter $\boldsymbol{\theta}$ and the graph signal $\mathbf{f}$ resembles the convolution of the standard Fourier analysis \cite{blackledge_chapter_2005}:
\begin{equation}\label{eq:gconv}
\mathcal{F}(\mathbf{f} \otimes \boldsymbol{\theta}) =  QWQ^T \mathbf{f}   
\end{equation}
where $\otimes$ is the convolution operator and $W = Q^T \boldsymbol{\theta}$ is a vector of learnable parameters. Repeated application of this convolution interleaved with nonlinearities led to the Spectral Convolutional Neural Network \cite{bruna_spectral_2014}.

This approach has some drawbacks. For instance, the parameters cannot be used for graphs with a different Laplacian, and their size grows linearly with the number of nodes in the graph. This, along with the need for an eigendecomposition, makes it difficult to deal with large graphs. Finally, the resulting filter may not be localized in space, \ie the filter does not modify the signal according to the neighborhood of each node only. This issues were later overcome \cite{defferrard_convolutional_2016} by using the truncated Chebyshev expansion \cite{hammond_wavelets_2011}. Interestingly, the Graph Convolutional Network \cite{kipf_semi-supervised_2017} layer truncates such expansion to the very first term, \ie the Laplacian of the graph, such that the node states at layer $\ell+1$ are computed (in matrix notation) as $H^{\ell+1} = \sigma(L{H^{\ell}}W)$, where $W$ is the matrix of learnable parameters. Therefore, this model represents an interesting connection between the local and iterative scheme of DGNs and spectral theory.

\subsection{Random-walks} In an attempt to capture local and global properties of the graph, random walks are often used to create node embeddings, and they have 
have been studied for a long time \cite{lovasz_random_1993,vishwanathan_graph_2010,ribeiro_struc2vec_2017,ivanov_anonymous_2018}. A random walk is defined as a random path that connects two nodes in the graphs. Depending on the reachable nodes, we can devise different frameworks to learn a node representation: for example, Node2Vec \cite{grover_node2vec_2016} maximizes the likelihood of a node given its surroundings by exploring the graph using a random walk. Moreover, learnable parameters guide the bias of the walk in the sense that a depth-first search can be preferred to a breadth-first search and vice-versa. Similarly, DeepWalk \cite{perozzi_deepwalk_2014} learns continuous node representations by modeling random walks as sentences and maximizing a likelihood objective. More recently, random walks have been used to generate graphs as well \cite{bojchevski_netgan_2018}, and a formal connection between the contextual information diffusion of GCN and random walks has been explored \cite{xu_representation_2018}.

\subsection{Adversarial training and attacks on graphs}
Given the importance of real-world applications that use graph data structures, there has recently been an increasing interest in studying the robustness of DGNs to malicious attacks. The term \emph{adversarial training} is used in the context of deep neural networks to identify a regularization strategy based on feeding the model with perturbed input. The catch is to make the network resilient to \emph{adversarial attacks} \cite{biggio_wild_2018}. Recently, neural DGNs have been shown to be prone to adversarial attacks as well \cite{zugner_adversarial_2018}, while the use of adversarial training for regularization is relatively new \cite{feng_graph_2019}. The adversarial training objective function is formulated as a min-max game where one tries to minimize the harmful effect of an adversarial example. Briefly, the model is trained with original graphs from the training set, as well as with adversarial graphs. Examples of perturbations to make a graph adversarial include arc insertion and deletions \cite{yang_topology_2019} or the addition of adversarial noise to the node representations \cite{jin_latent_2019}. The adversarial graphs are labeled according to their closest match in the dataset. This way, the space of the loss function is smooth, and it preserves the predictive power of the model even in the presence of perturbed graphs.

\subsection{Sequential generative models of graphs}
Another viable option to generate graphs is to model the generative process as a sequence of actions. This approach has been shown to be able to generalize to graphs coming from very different training distributions; however, it relies on a fixed ordering of graph nodes. A seminal approach is the one in \cite{li_learning_2018}, where the generation of a graph is modeled as a decision process. Specifically, a stack of neural networks is trained jointly to learn whether to add new nodes, whether to add new edges and which node to focus on the next iteration. Another work of interest is \cite{you_graphrnn_2018}, where the generation is formulated as an auto-regressive process where nodes are added sequentially to the existing graph. Each time a new node is added, its adjacency vector with respect to the existing nodes is predicted by a \quotes{node-level} Gated Recurrent Unit network \citep{kyunghyun_learning_2014}. At the same time, another \quotes{graph-level} network keeps track of the state of the whole graph to condition the generation of the adjacency vector. Finally, \cite{bacciu_graph_2019, bacciu_edge-based_2019} models the generative tasks by learning to predict the ordered edge set of a graph using two Gated Recurrent Unit networks; the first one generates the first endpoints of the edges, while the second predicts the missing endpoints conditioned on such information.
\section{Open Challenges and Research Avenues}\label{sec:open}
Despite the steady increase in the number of works on graph learning methodologies, there are some lines of research that have not been widely investigated yet. Below, we mention some of them to give practitioners insights about potential research avenues.

\subsection{Time-evolving graphs}
Current research has been mostly focusing on methods that automatically extract features from static graphs. However, being able to model dynamically changing graphs constitutes a further generalization of the techniques discussed in this survey. There already are some supervised \cite{li_gated_2016, wang_dynamic_2019} and unsupervised \cite{zambon_concept_2018} proposals in the literature. However, the limiting factor for the development of this research line seems, currently, the lack of large datasets, especially of non-synthetic nature. 

\subsection{Bias-variance trade-offs}
The different node aggregation mechanisms described in Section \ref{sec:neighborhood-aggregation} play a crucial role in determining the kind of structures that a model can discriminate. For instance, it has been proven that Graph Isomorphism Network is theoretically as powerful as the 1-dim Weisfeiler Lehman test of graph isomorphism \cite{xu_how_2019}. As a result, this model is able to overfit most of the datasets it is applied to. Despite this flexibility, it may be difficult to learn a function that generalizes well: this is a consequence of the usual bias-variance trade-off \cite{friedman_elements_2001}. Therefore, there is a need to characterize all node aggregation techniques in terms of structural discrimination power. A more principled definition of DGNs is essential to be able to choose the right model for a specific application.

\subsection{A sensible use of edge information}
Edges are usually treated as second-class citizens when it comes to information sources; indeed, most of the models which deal with additional edge features \cite{micheli_neural_2009, simonovsky_dynamic_2017, bacciu_contextual_2018, schlichtkrull_modeling_2018} compute a weighted aggregation where the weight is given by a suitable transformation of edge information. However, there are interesting questions that have not been answered yet. For example, is it reasonable to apply context spreading techniques to edges as well? The advantages of such an approach are still not clear. Furthermore, it would be interesting to characterize the discriminative power of methods that exploit edge information. 

\subsection{Hypergraph learning}
Hypergraphs are a generalization of graphs in which an edge is connected to a subset of nodes rather than just two of them. Some works on learning from hypergraph have recently been published \cite{zhou_learning_2007, zhang_dynamic_2018, feng_hypergraph_2019, jiang_dynamic_2019}, and the most recent ones take inspiration from local and iterative processing of graphs. Like time-evolving graphs, the scarce availability of benchmarking datasets makes it difficult to evaluate these methods empirically.
\section{Applications}\label{sec:applic}
Here, we give some examples of domains in which graph learning can be applied. We want to stress that the application of more general methodologies to problems that have been usually tackled by using flat or sequential representations may bring performance benefits. As graphs are ubiquitous in nature, the following list is far from being exhaustive. Nonetheless, we summarize some of the most common applications to give the reader an introductory overview. As part of our contribution, we release a software library that can be used to easily perform rigorous experiments with DGNs\footnote{The code is available at \url{https://github.com/diningphil/PyDGN}}.

\subsection{Chemistry and Drug Design}
Cheminformatics is perhaps the prominent domain where DGNs have been applied with success, and chemical compound datasets are often used to benchmark new models. At a high level, predictive tasks in this field concern learning a direct mapping between molecular structures and outcomes of interest. For example, the Quantitative Structure-Activity Relationship (QSAR) analysis deals with the prediction of the biological activity of chemical compounds. Similarly, the Quantitative Structure-Property Relationship (QSPR) analysis focuses on the prediction of  chemical properties such as toxicity and solubility. Instances of pioneering applications of models for structured data to QSAR/QSPR analysis are in \cite{bianucci_application_2000}, and see \cite{micheli_introduction_2007} for a survey. DNGNs have also been applied to the task of finding structural similarities among compounds \cite{duvenaud_convolutional_2015, jeon_fp2vec_2019}. Another interesting line of research is computational drug design, \eg drug side-effect identification \cite{zitnik_modeling_2018} and drug discovery. As regards the latter task, several approaches use deep generative models to discover novel compounds. These models also provide mechanisms to search for molecules with a desired set of chemical properties \cite{jin_junction_2018, samanta_nevae_2019, liu_constrained_2018}. In terms of benchmarks for graph classification, there is a consistent number of chemical datasets used to evaluate performances of DGNs. Among them, we mention NCI1 \cite{nci1}, PROTEINS, \cite{proteins} D\&D \cite{dd}, MUTAG \cite{mutag}, PTC \cite{ptc} and ENZYMES \cite{enzymes}. 

\subsection{Social networks}
Social graphs represents users as nodes and relations such as friendship or co-authorship as arcs. User representations are of great interest in a variety of tasks, for example to detect whether an actor in the graph is the potential source of misinformation or unhealthy behavior \cite{mishra_neural_2018, nechaev_sociallink_2018}. For these reasons, social networks are arguably the richest source of information for graph learning methods, in that a vast amount of features are available for each user. At the same time, the exploitation of this kind of information raises privacy and ethical concerns, this being the reason why datasets are not publicly available. The vast majority of supervised tasks on social graphs regards node and graph classification. In node classification, three major datasets in literature are usually employed to assess the performances of DGNs, namely Cora, Citeseer \cite{citeseer_cora} and  PubMed \cite{pubmed}, for which a rigorous evaluation is presented in \cite{shchur_pitfalls_2018}.
Instead, the most popular social benchmarks for (binary and multiclass) graph classification are IMDB-BINARY, IMDB-MULTI, REDDIT-BINARY, REDDIT-MULTI and COLLAB \cite{yanardag_deep_2015}. The results of a rigorous evaluation of several DGNs on these datasets, where graphs use uninformative node features, can be found in \cite{errica_fair_2020}.  

\subsection{Natural Language Processing}
Another interesting application field leverages graph learning methods for Natural Language Processing tasks, where the input is usually represented as a sequence of tokens. By means of dependency parsers, we can augment the input as a tree \cite{bacciu_deep_2020} or as a graph and learn a model that takes into account the syntactic \cite{marcheggiani_encoding_2017} and semantic \cite{marcheggiani_exploiting_2018} relations between tokens in the text. An example is neural machine translation, which can be formulated as a graph-to-sequence problem \cite{beck_graph--sequence_2018} to consider syntactic dependencies in the source and target sentence.

\subsection{Security}
The field of static code analysis is a promising new application avenue for graph learning methods. Practical applications include: i) determining if two assembly programs, which stem from the same source code, have been compiled by means of different optimization techniques; ii) prediction of specific types of bugs by means of augmented Abstract Syntax Trees \cite{iadarola_graph-based_2018}; iii) predicting whether a program is likely to be the \textit{obfuscated} version of another one; iv) automatically extracting features from Control Flow Graphs \cite{massarelli_safe_2019}.

\subsection{Spatio-temporal forecasting}
DGNs are also interesting to solve tasks where the structure of a graph changes over time. In this context, one is interested not only in capturing the structural dependencies between nodes but also in the evolution of these dependencies on the temporal domain. Approaches to this problem usually combine a DGN (to extract structural properties of the graph) and a Recurrent Neural Network (to model the temporal dependencies). Example of applications include the prediction of traffic in road networks \cite{yu_spatio-temporal_2018}, action recognition \cite{wang_videos_2018} and supply chain \cite{san_kim_graph_2019} tasks.

\subsection{Recommender Systems}
In the Recommender Systems domain \cite{bobadilla_recommender_2013}, graphs are a natural candidate to encode the relations between users and items to recommend. For example, the typical user-item matrix can be thought of as a bipartite graph, while user-user and item-item matrices can be represented as standard undirected graphs. Recommending an item to a user is a \quotes{matrix completion} task, \ie learning to fill the unknown entries of the user-item matrix, which can be equivalently formulated as a link prediction task. Based on these analogies, several DGN models have been recently developed to learn Recommender Systems from graph data \cite{monti_geometric_2017, yin_deeper_2019}. Currently, the main issues pertain scalability of computation to large graphs. As a result, techniques like neighborhood sampling have been proposed in order to reduce the computational overhead \cite{ying_graph_2018}.
\section{Conclusions}
After a pioneering phase in the early years of the millennia, the topic of neural networks for graph processing is now a consolidated and vibrant research area. In this expansive phase, research works at a fast pace producing a plethora of models and variants thereof, with less focus on systematization and tracking of early and recent literature. For the field to move further to a maturity phase, we believe that certain aspects should be deepened and pursued with higher priority. A first challenge, in this sense, pertains to a formalization of the different adaptive graph processing models under a unified framework that highlights their similarities, differences, and novelties. Such a framework should also allow reasoning on theoretical and expressiveness properties \cite{xu_how_2019} of the models at a higher level. A notable attempt in this sense has been made by \cite{gilmer_neural_2017}, but it does not account for the most recent developments and the variety of mechanisms being published (\eg pooling operators and graph generation, to name a few). An excellent reference, with respect to this goal, is the seminal work of \cite{frasconi_general_1998}, which provided a general framework for tree-structured data processing. This framework is expressive enough to generalize supervised learning to tree-to-tree non-isomorph transductions, and it generated a followup of theoretical research \cite{hammer_recursive_2004,hammer_universal_2005} which consolidated the field of recursive neural networks. The second challenge relates to the definition of a set of rich and robust benchmarks to test and assess models in fair, consistent, and reproducible conditions. Some works \cite{errica_fair_2020,shchur_pitfalls_2018} are already bringing to the attention of the community some troubling trends and pitfalls as concerns datasets and methodologies used to assess DGNs in the literature. We believe such criticisms should be positively embraced by the community to pursue the growth of the field. Some attempts to provide a set of standardized data and methods appear now under development\footnote{Open Graph Benchmark: \url{http://ogb.stanford.edu/}}. Also, recent progress has been facilitated by the growth and wide adoption by the community of new software packages for the adaptive processing of graphs. In particular, the PyTorch Geometrics \cite{fey_fast_2019} and Deep Graph Library \cite{wang_deep_2019} packages provide standardized interfaces to operate on graphs for ease of development. Moreover, they allow training models using all the Deep Learning tricks of the trade, such as GPU compatibility and graph mini-batching. The last challenge relates to applications. We believe a methodology reaches its maturity when it will show the transfer of research knowledge to an impactful innovation for the society. Again, attempts in this sense are already underway, with good candidates being in the fields of chemistry \cite{bradshaw_model_2019} and life-sciences \cite{zitnik_modeling_2018}.

\section*{Acknowledgements}
This work has been partially supported by the Italian Ministry of Education, University, and Research (MIUR) under project SIR 2014 LIST-IT (grant n. RBSI14STDE).

\newpage
\appendix
\section{Acronyms Table}
\label{appendix:A}

\begin{table}[ht]
\centering
\footnotesize
\renewcommand{\arraystretch}{2.0}
\begin{tabular}{lcc}
Acronym & Model Name & Reference  \\ \hline
GNN & Graph Neural Network & \cite{scarselli_graph_2009} \\
NN4G & Neural Network for Graphs & \cite{micheli_neural_2009} \\
GraphESN & Graph Echo State Network & \cite{gallicchio_graph_2010} \\
GCN & Graph Convolutional Network & \cite{kipf_semi-supervised_2017} \\
GG-NN & Gated Graph Neural Network & \cite{li_gated_2016} \\
ECC & Edge-Conditioned Convolution & \cite{simonovsky_dynamic_2017} \\
GraphSAGE & Graph SAmple and aggreGatE & \cite{hamilton_inductive_2017} \\
CGMM & Contextual Graph Markov Model & \cite{bacciu_contextual_2018} \\
DGCNN & Deep Graph Convolutional Neural Network & \cite{zhang_end--end_2018} \\    
DiffPool & Differentiable Pooling & \cite{ying_hierarchical_2018} \\
GAT & Graph Attention Network & \cite{velickovic_graph_2018} \\                     
R-GCN & Relational Graph Convolutional Network & \cite{schlichtkrull_modeling_2018} \\
DGI & Deep Graph Infomax & \cite{velickovic_deep_2019} \\
GMNN & Graph Markov Neural Network & \cite{qu_gmnn_2019} \\
GIN & Graph Isomorphism Network & \cite{xu_how_2019} \\
NMFPool & Non-Negative Matrix Factorization Pooling &  \cite{bacciu_non-negative_2019}\\
SAGPool & Self-attention Graph Pooling & \cite{lee_self-attention_2019} \\
Top-k Pool & Graph U-net & \cite{gao_graph_2019} \\
FDGNN & Fast and Deep Graph Neural Network & \cite{gallicchio_fast_2020} \\ \\
\end{tabular}
\caption{Reference table with acronyms, their extended names, and associated references. }
\label{tab:acronyms-table}
\end{table}

\newpage
\bibliographystyle{plain}
\bibliography{main}

\end{document}